\documentclass[10pt,twocolumn,letterpaper]{article}

\usepackage[pagenumbers]{cvpr} 

\usepackage{multirow}
\usepackage{makecell} 
\usepackage{graphicx}
\PassOptionsToPackage{dvipsnames,table}{xcolor}
\usepackage{colortbl}
\usepackage{array}
\usepackage{textcomp}
\usepackage{paralist}
\usepackage[accsupp]{axessibility}  

\definecolor{cvprblue}{rgb}{0.21,0.49,0.74}
\usepackage[pagebackref,breaklinks,colorlinks,allcolors=cvprblue]{hyperref}

\def\paperID{5113}
\def\confName{CVPR}
\def\confYear{2025}

\title{SAM-I2V: Upgrading SAM to Support Promptable Video Segmentation\\with Less than 0.2\% Training Cost}

\author{
Haiyang Mei,\hspace{0.12cm}
Pengyu Zhang,\hspace{0.12cm}
Mike Zheng Shou*\\
Show Lab, National University of Singapore
}

\begin{document}
\maketitle

\renewcommand{\thefootnote}{}
\footnotetext{*Corresponding Author}

\begin{abstract}
Foundation models like the Segment Anything Model (SAM) have significantly advanced promptable image segmentation in computer vision. However, extending these capabilities to videos presents substantial challenges, particularly in ensuring precise and temporally consistent mask propagation in dynamic scenes. SAM 2 attempts to address this by training a model on massive image and video data from scratch to learn complex spatiotemporal associations, resulting in huge training costs that hinder research and practical deployment. In this paper, we introduce SAM-I2V, an effective image-to-video upgradation method for cultivating a promptable video segmentation (PVS) model. Our approach strategically upgrades the pre-trained SAM to support PVS, significantly reducing training complexity and resource requirements. To achieve this, we introduce three key innovations: (i) an image-to-video feature extraction upgrader built upon SAM’s static image encoder to enable spatiotemporal video perception, (ii) a memory filtering strategy that selects the most relevant past frames for more effective utilization of historical information, and (iii) a memory-as-prompt mechanism leveraging object memory to ensure temporally consistent mask propagation in dynamic scenes. Comprehensive experiments demonstrate that our method achieves over 90\% of SAM 2's performance while using only 0.2\% of its training cost. Our work presents a resource-efficient pathway to PVS, lowering barriers for further research in PVS model design and enabling broader applications and advancements in the field. Code and model are available at: \url{https://github.com/showlab/SAM-I2V}.
\end{abstract}
    
\section{Introduction}
\label{sec:introduction}
Foundation models have revolutionized artificial intelligence by leveraging massive datasets and large-scale models to achieve remarkable performance across various tasks. In computer vision, the Segment Anything Model (SAM)~\cite{sam1} has emerged as a prominent foundation model for general-purpose promptable image segmentation, enabling users to segment objects/regions in images based on prompts such as points, boxes, or masks. While SAM demonstrates impressive capabilities on static images, extending these abilities to video data introduces significant challenges due to the temporal dynamics and consistency required in video segmentation tasks.

\begin{figure}[t]
    \centering
    \includegraphics[width=0.99\linewidth]{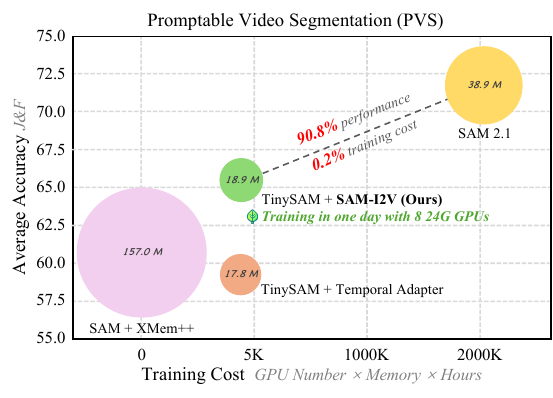}
    \vspace{-8pt}
    \caption{Comparison on promptable video segmentation (PVS). We depict the online-PVS performance (in \textit{J\&F}) averaged on four benchmark datasets \cite{huang2023neuromorphic,bekuzarov2023xmem++,wang2023towards,sam2} (vertical axis) with respect to model training cost (horizontal axis). The size of circles indicates model parameters (in M). SAM 2.1 \cite{sam2} is a strong PVS model but needs huge training cost (\ie, 256 A100 GPUs for 108 hours). We propose to cultivate a PVS model via image-to-video model upgradation. Building on the pre-trained SAM \cite{sam1, shu2023tinysam}, incorporating existing video object segmentation models (\eg, XMem++ \cite{bekuzarov2023xmem++}) or temporal association modules (\eg, temporal adapter \cite{sam2}) significantly reduces training costs compared to SAM 2.1 \cite{sam2}. Furthermore, our designed SAM-I2V helps achieve over 90\% of SAM 2.1's performance while using only 0.2\% of its training cost.}
    \label{fig:teaser}
    \vspace{-10pt}
\end{figure}

Recently, SAM 2~\cite{sam2} has been developed to address promptable video segmentation (PVS), aiming to extend SAM's image segmentation capabilities to videos. SAM 2 is trained from scratch on a massive collection of images and videos to learn complex spatiotemporal associations. However, it demands exorbitant computational resources, requiring 256 A100 GPUs over 108 hours for training. Such high computational costs create substantial barriers to exploring alternative architectures and advancing research in PVS, limiting accessibility for researchers and practitioners.

This raises a critical question: \textit{Can we achieve promptable video segmentation through a resource-efficient approach?} If so, it would open opportunities for developing diverse PVS foundation models, benefiting a wide range of downstream applications. Addressing this challenge, we propose \textbf{SAM-I2V}, an efficient method to upgrade the pre-trained image-based SAM~\cite{sam1} model to support promptable video segmentation with affordable training cost (\autoref{fig:teaser}). Our approach strategically leverages SAM's strong promptable image segmentation capabilities and develops image-to-video upgradation techniques to extend functionality from images to videos.

To bridge the gap between promptable image and video segmentation, our SAM-I2V includes three key components: a temporal feature integrator TFI, a memory selective associator MSA, and a memory prompt generator MPG. (i) The TFI enhances SAM's image encoder by integrating temporal information, enabling the extraction of spatiotemporal features essential for understanding dynamic scenes; (ii) The MSA retrieves relevant segmentation information from preceding video frames through a new memory filtering mechanism, selectively associating historical information to guide the current frame's segmentation; (iii) The MPG incorporates memory as an additional prompt, utilizing object memory to ensure precise and temporally consistent mask propagation across frames.
Comprehensive experiments validate that our method significantly reduces training complexity and resource requirements, achieving over $90\%$ of SAM 2's performance with only $0.2\%$ of its training cost.
In summary, our contributions are:
\begin{compactenum}
    \item We present SAM-I2V, the first attempt to upgrade a pre-trained image-based SAM model to support promptable video segmentation, enabling PVS capabilities with substantially reduced training costs.
    \item We design an upgrade method that extends SAM’s static image encoder into a video perception module, enabling spatiotemporal feature extraction for PVS.
    \item We propose a memory filtering strategy that selects the most relevant past frames to facilitate more effective utilization of historical information for memory-guided video segmentation.
    \item We introduce a memory-as-prompt mechanism that leverages object memory to ensure temporally consistent mask propagation in dynamic scenes.
\end{compactenum}

\section{Related Works}
\label{related_work}
\textbf{Promptable Visual Segmentation.}
The Segment Anything Model (SAM) \cite{sam1} has set a significant milestone in promptable segmentation. SAM is designed to generate a valid segmentation mask for an object of interest in an image when provided with input prompts such as bounding boxes or reference points. Trained on the extensive SA-1B dataset, SAM facilitates zero-shot segmentation with its versatile prompting mechanism, making it highly adaptable for diverse downstream tasks. This foundational work has spurred further innovations aimed at enhancing SAM’s quality and efficiency. For instance, models like HQ-SAM \cite{ke2024segment} have focused on improving output precision through high-quality token integration and fine-tuning with detailed masks. Efficiency-oriented models such as EfficientSAM \cite{xiong2024efficientsam}, MobileSAM \cite{zhang2023faster,zhang2023mobilesamv2}, FastSAM \cite{zhao2023fast}, TinySAM \cite{shu2023tinysam}, and SlimSAM \cite{chen20230slimsam} have broadened the scope of SAM’s applications, enabling its deployment in more resource-constrained and real-world scenarios, including medical imaging \cite{ma2024segment,mazurowski2023segment,wu2023medical} and remote sensing \cite{chen2024rsprompter,ren2024segment,catsam}.
Building upon the success of SAM \cite{sam1}, SAM 2 \cite{sam2} represents an evolution toward promptable video segmentation, addressing the complexities of propagating segmentation across multiple frames. This allows for interaction at any frame of a video, accepting prompts such as positive/negative clicks, bounding boxes, or masks to define or refine object segmentation. Unlike its predecessor, SAM 2 not only responds to initial prompts on single frames but also propagates these segmentation cues throughout the entire video, maintaining temporal consistency and adaptability across frames. Evaluated through interactive online and offline settings, SAM 2's integration of promptable video interaction ensures comprehensive and dynamic segmentation, pushing the boundaries of practical real-world applicability \cite{dong2024segment,zhu2024medical,shen2024interactive,liu2024surgical,xiong2024sam2,chen2024sam2}.

\noindent\textbf{Interactive Video Object Segmentation.}
Interactive video object segmentation (IVOS) is an essential task for enabling real-time object tracking and segmentation in videos through user input, such as clicks or bounding boxes. Recent studies \cite{heo2020interactive,cheng2021modular,delatolas2024learning} have aimed to develop modular approaches that allow segmenting an object in a single frame and then propagating this segmentation across subsequent frames to maintain consistency. Building upon this, many IVOS methods leverage interactive mechanisms to guide segmentation over multiple frames. While SAM has been used for video object segmentation combined with video trackers based on masks \cite{cheng2023tracking,yang2023track,cheng2023segment} or points \cite{rajivc2023segment}, these approaches face certain limitations: trackers may not be effective for all objects, SAM may perform suboptimally on certain video frames, and there is no interactive mechanism for correcting model errors aside from re-annotating erroneous frames and restarting the tracking process from that point. This highlights expanded functionality of SAM 2 that provides a flexible solution to enable interactive refinement.

\noindent\textbf{Semi-Supervised Video Object Segmentation.}
Semi-supervised video object segmentation (Semi-VOS) typically begins with a provided mask for the first frame, which acts as a reference for tracking the object through the video \cite{pont20172017}. The term “semi-supervised” refers to using the initial mask as supervision, with no further manual input for subsequent frames. This task has become vital for various applications requiring automatic tracking, such as video editing, robotics, and automatic background removal.
Early Semi-VOS techniques involved fine-tuning models on the first frame \cite{caelles2017one,maninis2018video,robinson2020learning} or incorporating offline-trained models that predicted object masks for the entire video \cite{oh2018fast,yang2018efficient,yang2020collaborative}. More advanced strategies integrated RNNs \cite{xu2018youtube} and cross-attention \cite{cheng2021rethinking,cheng2022xmem,bekuzarov2023xmem++,cheng2024putting} to extend segmentation from the first frame throughout the sequence, improving object tracking accuracy. Recent methods \cite{zhang2023joint,wu2023scalable} have enhanced a single vision transformers to jointly handle the current frame alongside all prior frames and their corresponding predictions, simplifyiing the architecture but incuring a significant inference cost. Semi-VOS process is analogous to a specialized case of promptable video segmentation, where providing an high-quality mask prompt in the first frame is crucial but practically challenging and time-consuming, limiting the practical application of Semi-VOS in real-world scenarios.

\begin{figure*}[ht]
    \centering
    \includegraphics[width=1\linewidth]{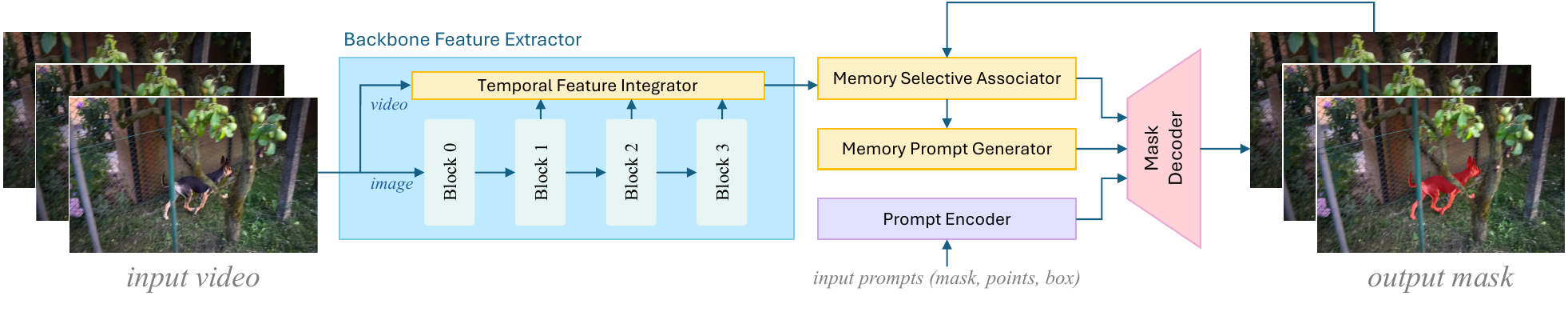}
    \vspace{-0.7cm}
    \caption{Overview of SAM-I2V. The input video is fed into a backbone feature extractor, which passes the current frame through a series of image encoder layers. The output of these layers is enhanced with temporal information by a temporal feature integrator to capture dynamic context across frames. The extracted features are then fed into the memory associator and the memory prompt generator, which manage memory states and generate target object prompts based on past frame information. The prompt encoder processes optional user input (\textit{e.g.}, masks, points, or bounding boxes) to guide the segmentation. Finally, the processed information flows into the mask decoder, which outputs the segmentation mask for current frame, enabling user-intention-followed or memory conditioned promptable video segmentation.}
    \vspace{-0.3cm}
    \label{fig:pipeline}
\end{figure*}

\noindent\textbf{Image-to-Video Transfer Learning.}
In recent years, the importance of image-to-video transfer learning has grown due to its significant potential for reducing training costs and enhancing adaptability in video-based tasks. To facilitate efficient image-to-video transfer, various methods have been introduced, including spatio-temporal adapters \cite{pan2022st,yang2023aim}, dual-path adaptation \cite{park2023dual}, disentangled spatial and temporal learning \cite{qing2023disentangling}, and learnable prompts \cite{ju2022prompting,lin2022frozen}. These methods incorporate lightweight, trainable components that leverage pre-trained image foundation models such as ViT \cite{dosovitskiy2020image} and CLIP \cite{radford2021learning} while integrating spatio-temporal dynamics, striking a balance between computational efficiency and performance to enhance video applications.
These methods, although centered around image-to-video transfer for video recognition and understanding tasks, offer insights into our efficient cultivation of a promtable video segmentation model (\ie, SAM 2 \cite{sam2}) via upgrading the pre-trained image-based SAM model \cite{sam1} to a video one. These approaches provide a solid foundation for spatio-temporal information learning in our image-to-video upgradation but lack direct applicability for addressing segmentation-specific challenge, \ie, ensuring precise and temporally consistent mask propagation in scene dynamics. This makes our focus on SAM to SAM 2 upgradation a distinct and challenging endeavor.

\section{Methodology}
\label{methodology}
While SAM 2 demonstrates strong capabilities, its exorbitant training demands—requiring 256 A100 GPUs over 108 hours—present a significant barrier to exploring alternative architectures and advancing research in PVS. To overcome this challenge, we aim to develop an efficient method for cultivating a PVS model with affordable training costs. Our strategy leverages the pre-trained SAM model, which already performs well on single frames, and upgrades it to SAM 2. We design three components to achieve this transformation: a Temporal Feature Integrator (Section \ref{sec:tfi}), a Memory Associator (\ref{sec:msa}), and a Memory Prompt Generator (\ref{sec:mpg}), which collectively upgrade image-based SAM into a promptable video segmentation model.

\subsection{Overview}
Our SAM-I2V enables image-based SAM model can process video input for promtable video segmentation. As illustrated in \autoref{fig:pipeline}, the video frames are passed through a backbone feature extractor, which processes the current frame using a series of image encoder layers. These encoder outputs are enriched with temporal context through a temporal feature integrator, enabling the capture of dynamic information across consecutive frames. The enhanced features are subsequently input into the memory selective associator and memory prompt generator modules. These components manage memory states and generate target object prompts informed by past frames data, ensuring continuity and contextual awareness throughout the video. Optional user-provided prompts (\eg, masks, points, or bounding boxes) are processed by the prompt encoder to guide the segmentation with user-specific inputs. Finally, the integrated information is directed into the mask decoder, which produces the final segmentation mask for the current frame.

\subsection{Temporal Feature Integrator}
\label{sec:tfi}
To enhance SAM's image encoder, which is limited to extracting only spatial features, into a video perception module, we introduce a Temporal Feature Integrator (TFI). The TFI comprises a temporal branch and an integration branch that operate synergistically to iteratively extract and integrate temporal features, enabling a comprehensive spatio-temporal feature extraction.

\begin{figure}[ht]
    \centering
    \includegraphics[width=1\linewidth]{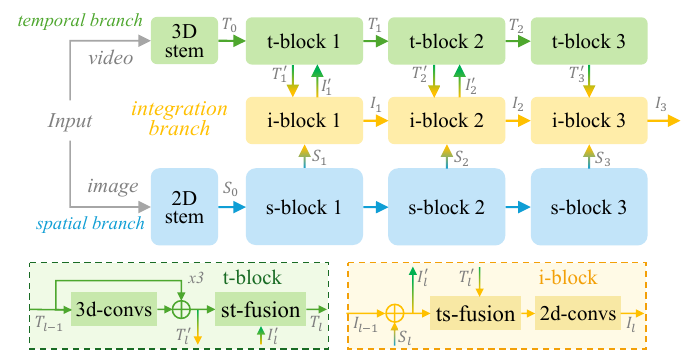}
    \caption{Structural details of temporal feature integrator (TFI). It consists of a temporal branch and an integration branch, enhancing SAM’s image encoder to function as a video perceptor.}
    \vspace{-0.3cm}
    \label{fig:tfi}
\end{figure}

As illustrated in \autoref{fig:tfi}, given the input video $X \in \mathbb{R}^{c \times n \times h \times w}$, where $c$, $n$, $h$ and $w$ are the frame channel, number, height and width, respectively, SAM's image encoder processes the current frame $X_t$ via a 2D stem and $L$ spatial blocks, generating multi-stage spatial features $S_l\in \mathbb{R}^{c_l \times h_l \times w_l}$, where $c_l$, $h_l$ and $w_l$ denote the features channel, height and width at stage $l\in\{1,\cdots,L\}$, respectively. The extracted features are fed into our TFI to obtain spatiotemporal features $I_l$.

Specifically, the temporal branch of TFI processes the video input $X$ through a 3D stem followed by $L$ temporal blocks to capture temporal contexts. In $l$-th temporal block, we employ 3D convolutions $\mathcal{C}_{\textit{3d}}$ to progressively model temporal correspondences between frames:
\[
\begin{aligned}
T'_l &= T^{(3)}, \\
T^{(i)} &= T^{(i-1)} + \mathcal{C}_{\textit{3d}}^{(i)}\big(T^{(i-1)}\big), \\
\end{aligned}
\]
where $T^{(0)} = T_{l-1}$ is the input to the $l$-th temporal block. To further enrich temporal features \(T'_l\) with spatial details, we fuse it with intermediate features \(I'_l\) from TFI's integration branch:
\[
T_l = \mathcal{ST}(I'_l, T'_l),
\]
where \(\mathcal{ST}\) denotes the spatial-to-temporal fusion operation that slices out current-frame temporal features from \(T'_l\) (\ie, the last slice in temporal dimension), concatenates it with current-frame spatial features \(I'_l\) along the channel dimension, applies a \(1 \times 1\) convolution to integrate concatenated features, and uses the resulting output to overwrite original current-frame temporal features in \(T'_l\).

Then, TFI's integration branch combines spatial features \( S_l \) and temporal features \( T'_l \) to get spatiotemporal features:
\[
\begin{aligned}
I'_l &= I_{l-1} + S_l, \\
I_l &= \mathcal{C}_{\textit{2d}}\big( \mathcal{TS}(I'_l, T'_l) \big),
\end{aligned}
\]
where \( I_{l-1} \) is the input to the \( l \)-th integration block. The temporal-to-spatial fusion operation \( \mathcal{TS} \) merges current-frame temporal features in \( T'_l \) with \( I'_l \), producing temporal-enriched spatial features. These features then pass through a 2D convolution \( \mathcal{C}_{\textit{2d}} \), yielding the enhanced spatiotemporal output \( I_l \).
Finally, the temporal and integration branches of TFI iteratively refine the features over multiple stages, enabling the model to progressively capture rich spatiotemporal context for the subsequent segmentation process.

\subsection{Memory Selective Associator}
\label{sec:msa}
While image-based SAM \cite{sam1} segments the target object based on user prompts, in the PVS task, segmentations from preceding video frames can serve as target indicators to guide the segmentation of the target in the current frame. This motivates our Memory Selective Associator (MSA), designed to retrieve relevant and beneficial segmentation information from preceding video frames, leveraging this memory to enhance the current frame's features.

\begin{figure*}[ht]
    \centering
    \includegraphics[width=1\linewidth]{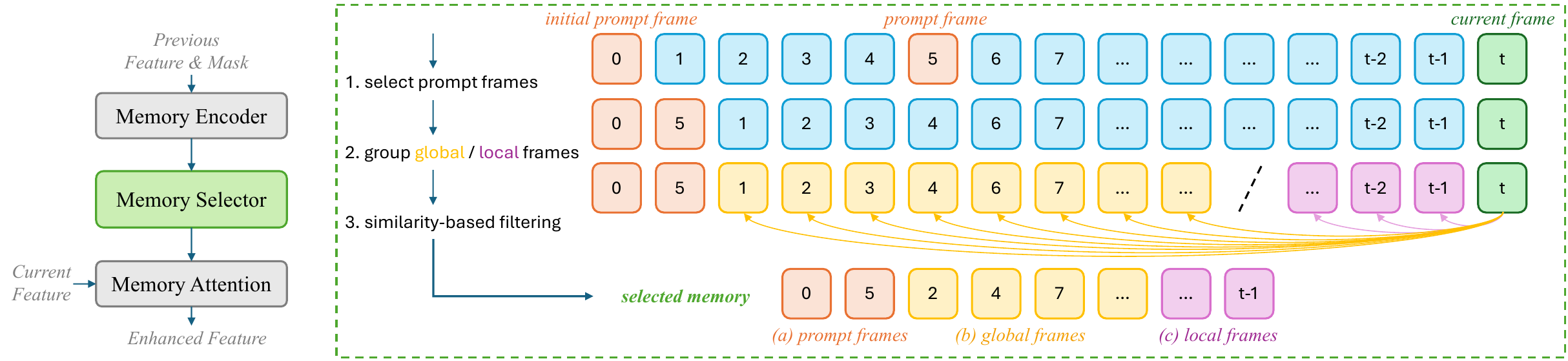}
    \caption{Illustration of our memory selective associator (MSA). It consists of a memory encoder, a memory selector, and memory attention.}
    \vspace{-0.3cm}
    \label{fig:msa}
\end{figure*}

MSA first utilizes convolution to fuse the features from the previous frames \( \{F_{t-p}\}_{p=1}^P \) and their corresponding segmentation masks \( \{M_{t-p}\}_{p=1}^P \). These fused features are then passed to the memory selector, which chooses the most relevant past frame features. The selected features are used in the memory attention module, where cross-attention between the current frame features \( I_t \) and the selected previous frame features is applied, producing the memory-enhanced features \( I'_t \).

The memory encoder and attention mechanism in our MSA follow a similar structure as in SAM 2 \cite{sam2}. Unlike SAM 2 \cite{sam2}, which selects the previous \( z \) consecutive frames leading up to the current frame, our memory selector chooses \( z \) frames based on their relevance, as determined by similarity to the current frame. Specifically, it first chooses the prompt frame based on the user input, and then divides the previous frames into \( X \) short-term (local) frames and \( Y \) long-term (global) frames. The similarity between the current frame features \( I_t \in \mathbb{R}^{N} \) and \( X/Y \) previous frame features \( F \in \mathbb{R}^{X/Y \times N} \) in each set is computed to identify the most relevant local/global frames:
\[
S^{loc} = F^{loc} \cdot I_t \in \mathbb{R}^{X}, \quad S^{glo} = F^{glo} \cdot I_t \in \mathbb{R}^{Y},
\]
where \( N = c_L \times h_L \times w_L \).
Next, we normalize the similarity scores to obtain the probability distribution \( D^{loc} \) and \( D^{glo} \), respectively:
\[
D^{loc} = \frac{e^{S^{loc}}}{\sum_j e^{S_j^{loc}}} \in \mathbb{R}^{X}, \quad
D^{glo} = \frac{e^{S^{glo}}}{\sum_j e^{S_j^{glo}}} \in \mathbb{R}^{Y}.
\]
We then sample \( x \) local frames and \( y=z-x \) global frames based on \( D^{loc} \) and \( D^{glo} \):
\[
i_k^{loc} \sim P(i_k^{loc} = i) = D^{loc}(i), \quad k \in \{1, 2, \dots, X\},
\]
\[
i_k^{glo} \sim P(i_k^{glo} = i) = D^{glo}(i), \quad k \in \{1, 2, \dots, Y\},
\]
where \( i_k^{loc} \) and \( i_k^{glo} \) denote the indices of the selected local and global frames, respectively, and \( D^{loc}(i) \) and \( D^{glo}(i) \) are the corresponding probabilities for each frame. The selected frames are subsequently used to enhance the segmentation of the current frame through memory attention, resulting in memory-associated features \( I_t' \).

\subsection{Memory Prompt Generator}
\label{sec:mpg}
SAM's mask decoder segments images using image features and user prompts to achieve intention-followed results. Motivated by its prompt integration capability, we propose incorporating memory as an additional prompt. This enables the model to leverage segmentation information from previous frames, enhancing temporal consistency across the video sequence.

\begin{figure}[t]
    \centering
    \includegraphics[width=1\linewidth]{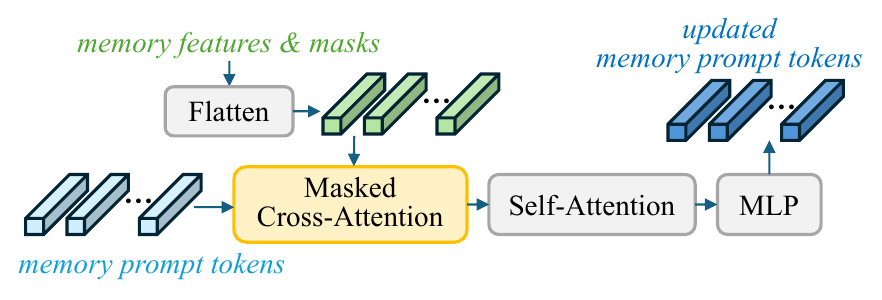}
    \vspace{-0.7cm}
    \caption{Illustration of the memory prompt generator (MPG).}
    \vspace{-0.2cm}
    \label{fig:mpg}
\end{figure}

\autoref{fig:mpg} illustrates the proposed Memory Prompt Generator (MPG). We introduce a set of learnable memory prompt tokens \( G \in \mathbb{R}^{g \times d} \), where \( g \) is the number of memory prompts and \( d \) is the embedding dimension. These tokens interact with the flattened memory features \( F_{\text{mem}} \in \mathbb{R}^{(m h_L w_L) \times c_L} \), where \( m \) is the number of memory frames, guided by the corresponding binarized segmentation masks \( M_{\text{mem}} \in \mathbb{R}^{(m h_L w_L) \times 1} \) via masked cross-attention:
\begin{align}
G' &= \text{Softmax}(\mathcal{M} + Q K^\top) V + G, \\
Q = \psi&(G), \quad K = \psi(F_{\text{mem}}), \quad V = \psi(F_{\text{mem}}), \\
\mathcal{M}(x, y) &=
\begin{cases}
0, & \text{if } M_{\text{mem}}(x, y) = 1, \\
-\infty, & \text{otherwise},
\end{cases}
\end{align}
where \( \psi \) is the learnable linear transformation. The updated memory prompt tokens \( G' \), having captured foreground context from memories, are then refined through self-attention to incorporate intra-token dependencies:
\[
G'' = \text{SelfAttn}(G').
\]
Finally, the refined memory prompt embeddings \( G'' \) are passed through a multi-layer perceptron (MLP) to produce the final updated memory prompt tokens:
\[
G''' = \text{MLP}(G'').
\]
Such an iterative attention-based token generation enables the MPG to selectively capture target object information from memories, providing guidance for the mask decoder.

\section{Experiments}\label{experiments}
\subsection{Experimental Setup}
\textbf{Training Datasets.}
We train our SAM-I2V jointly on 10,000 SA-1B \cite{sam1} images and 50,583 SAV \cite{sam2} videos.
The SA-1B dataset \cite{sam1} is a comprehensive collection consisting of over 1.1 billion high-quality segmentation masks from 11 million licensed and privacy-preserving images. This dataset was assembled using a multi-phase data engine where initial stages involved human annotation aided by interactive tools, and subsequent phases transitioned to fully automated mask generation through the SAM model. The rich variety of mask annotations allows for robust training that supports zero-shot generalization and diverse segmentation tasks, proving effective in many downstream applications.
The Segment Anything Video (SA-V) dataset \cite{sam2} complements the image-focused SA-1B by providing extensive coverage for video segmentation. Its training set includes 50,583 videos and 642,036 masklets. SA-V is tailored for video scenarios involving complex object motions, occlusions, and reappearances, making it invaluable for advancing video-based models like SAM 2 \cite{sam2}.

\noindent\textbf{Training Configurations.}
Following \cite{sam2}, to simulate an interactive training setting, we sample 8-frame sequences, designating the first frame as the prompt frame and randomly selecting up to 1 additional frame from the remaining seven frames for corrective clicks. The initial prompt for the first frame can be the ground-truth mask (50\% probability), a positive click from the ground-truth mask (25\%), or a bounding box input (25\%). The corrective clicks are sampled from the error region between the model prediction and the ground-truth mask (90\%) or directly from the ground-truth mask (10\%).
Our SAM-I2V model is implemented using PyTorch~\cite{paszke2019pytorch} and trained for 10 epochs on 8 RTX A5000 GPUs (24GB memory). We employ the AdamW optimizer~\cite{loshchilov2017decoupled} and apply layer-wise decay~\cite{clark2020electra} specifically to the image encoder. The initial learning rates are set to $3 \times 10^{-4}$ for the image encoder and $6 \times 10^{-5}$ for the other components, respectively. These learning rates are subsequently decayed using a cosine scheduler.
We supervise the model’s predictions using a linear combination of mask prediction loss for the output mask, mean-absolute-error (MAE) loss for the IoU prediction, and cross-entropy loss for object prediction with a ratio of 20:1:1, respectively. The mask prediction loss $\mathcal{L}_{m}$ is defined as the combination of the weighted binary cross-entropy (BCE) loss $\ell_{wbce}$ \cite{wei2019f3net} and the weighted IoU loss $\ell_{wiou}$ \cite{wei2019f3net}, \ie, $\mathcal{L}_{m} = \ell_{wbce} + \ell_{wiou}$, to ensure the model pays more attention to potential distraction regions such as object boundaries, elongated areas, or holes.
Notably, all SAM-I2V modules except the TinySAM \cite{shu2023tinysam} backbone are randomly initialized and trained from scratch without leveraging any weights from SAM 2 \cite{sam2}, ensuring complete independence from SAM 2 \cite{sam2} in the training process.

\begin{table*}[htbp]
    \centering
    \scriptsize
    \renewcommand{\arraystretch}{1.75}
    \setlength{\tabcolsep}{3.8pt} 
    \begin{tabular}{l|c|c|c|c|c|c|c|c|c|c|c|c|c|c|c|c|c|c|c}
        \hline
        \hline
        \multirow{2}{*}{Methods} & \multirow{2}{*}{\makecell{Para.\\(M)}} & \multirow{2}{*}{Cost} & \multicolumn{4}{c|}{1-click} & \multicolumn{4}{c|}{3-click} & \multicolumn{4}{c|}{5-click} & \multicolumn{4}{c|}{Average} & \multirow{2}{*}{OA} \\ \cline{4-19}
        & & & ES & PU & LV & SV & ES & PU & LV & SV & ES & PU & LV & SV & ES & PU & LV & SV & \\ 
        \hline
        \rowcolor{gray!7} SAM 2.1 \cite{sam2} & 38.9 & 2.2m & 63.6 & 45.8 & 59.3 & 66.3 & 86.8 & 66.8 & 79.5 & 75.2 & 88.8 & 70.1 & 82.8 & 77.2 & 79.7 & 60.9 & 74.9 & 72.9 & 71.9 \\ \hline
        \hline
        \rowcolor{orange!7} SAM \cite{sam1} + XMem++ \cite{bekuzarov2023xmem++} & 157.0 & - & 78.3 & 49.3 & 68.0 & 42.4 & 78.8 & 46.2 & 73.0 & 41.9 & 79.3 & 45.4 & 78.4 & 43.3 & 78.8 & 47.0 & 73.1 & 42.5 & 60.4 \\ \hline
        \rowcolor{orange!7} SAM \cite{sam1} + Cutie \cite{cheng2024putting} & 129.8 & - & 78.4 & 38.3 & 69.4 & 51.2 & 79.0 & 44.3 & 73.9 & 51.8 & 79.8 & 44.6 & 76.9 & 52.9 & 79.1 & 42.4 & 73.4 & 52.0 & 61.7 \\ \hline
        \rowcolor{yellow!7} TinySAM \cite{shu2023tinysam} + XMem++ \cite{bekuzarov2023xmem++} & 72.3 & - & 79.3 & 55.2 & 68.5 & 47.3 & 82.9 & 59.8 & 74.9 & 48.7 & 84.2 & 60.4 & 76.9 & 51.6 & 82.1 & 58.5 & 73.4 & 49.2 & 65.8 \\ \hline
        \rowcolor{yellow!7} TinySAM \cite{shu2023tinysam} + Cutie \cite{cheng2024putting} & 45.1 & - & 80.6 & 53.0 & 71.1 & 58.4 & 83.3 & 52.6 & 76.2 & 59.9 & 84.4 & 56.2 & 78.8 & 63.1 & 82.8 & 53.9 & 75.4 & 60.5 & 68.1 \\
        \hline
        \hline
        \rowcolor{pink!7} TinySAM \cite{shu2023tinysam} + TA \cite{sam2} & 17.8 & 4.6k & 68.7 & 28.8 & 56.2 & 42.9 & 80.2 & 45.1 & 72.3 & 49.1 & 82.1 & 49.8 & 76.5 & 50.9 & 77.0 & 41.2 & 68.4 & 47.6 & 58.6 \\ \hline
        \rowcolor{pink!7} TinySAM \cite{shu2023tinysam} + SA \cite{chen2022adaptformer} + TA \cite{sam2} & 18.7 & 4.6k & 54.8 & 34.6 & 52.8 & 44.1 & 79.3 & 52.7 & 72.9 & 50.6 & 81.9 & 59.4 & 76.8 & 51.9 & 72.0 & 48.9 & 67.5 & 48.9 & 59.3 \\ \hline
        \rowcolor{green!7} TinySAM \cite{shu2023tinysam} + \textbf{SAM-I2V} (Ours) & 18.9 & 4.6k & 73.0 & 39.1 & 53.8 & 50.5 & 84.7 & 60.8 & 73.4 & 58.2 & 86.4 & 65.9 & 78.3 & 59.5 & 81.4 & 55.3 & 68.5 & 56.1 & \textbf{65.3} \\ 
        \hline
        \hline
    \end{tabular}
    \caption{Comparisons of the online promptable video segmentation with 1-click, 3-click, and 5-click settings on four benchmark datasets (\ie, ES \cite{huang2023neuromorphic}, PU \cite{bekuzarov2023xmem++}, LV \cite{wang2023towards}, and SV \cite{sam2}). ``-'' indicates directly combining existing pre-trained models for inference. ``\textit{Cost}'' is calculated by \textit{GPU number} $\times$ \textit{GPU memory} $\times$ \textit{training hours}. ``\textit{OA}'' denotes the overall segmentation performance.}
    \label{tab:online}
\end{table*}

\begin{table*}[htbp]
    \centering
    \scriptsize
    \renewcommand{\arraystretch}{1.75}
    \setlength{\tabcolsep}{3.8pt} 
    \begin{tabular}{l|c|c|c|c|c|c|c|c|c|c|c|c|c|c|c|c|c|c|c}
        \hline
        \hline
        \multirow{2}{*}{Methods} & \multirow{2}{*}{\makecell{Para.\\(M)}} & \multirow{2}{*}{Cost} & \multicolumn{4}{c|}{1-click} & \multicolumn{4}{c|}{3-click} & \multicolumn{4}{c|}{5-click} & \multicolumn{4}{c|}{Average} & \multirow{2}{*}{OA} \\ \cline{4-19}
        & & & ES & PU & LV & SV & ES & PU & LV & SV & ES & PU & LV & SV & ES & PU & LV & SV & \\ 
        \hline
        \rowcolor{gray!7} SAM 2.1 \cite{sam2} & 38.9 & 2.2m & 63.9 & 46.2 & 61.3 & 66.8 & 87.7 & 67.3 & 83.9 & 76.2 & 90.1 & 70.7 & 87.5 & 78.1 & 80.6 & 61.4 & 77.6 & 73.7 & 73.3 \\ \hline
        \hline
        \rowcolor{orange!7} SAM \cite{sam1} + XMem++ \cite{bekuzarov2023xmem++} & 157.0 & - & 49.5 & 36.8 & 50.1 & 41.1 & 64.0 & 49.3 & 53.7 & 44.1 & 65.3 & 54.7 & 55.2 & 44.9 & 59.6 & 46.9 & 53.0 & 43.4 & 50.7 \\ \hline
        \rowcolor{yellow!7} TinySAM \cite{shu2023tinysam} + XMem++ \cite{bekuzarov2023xmem++} & 72.3 & - & 47.7 & 46.0 & 47.5 & 49.1 & 73.6 & 55.4 & 52.9 & 55.5 & 75.9 & 62.5 & 53.7 & 58.2 & 65.7 & 54.6 & 51.4 & 54.3 & 56.5 \\ \hline
        \hline
        \rowcolor{pink!7} TinySAM \cite{shu2023tinysam} + TA \cite{sam2} & 17.8 & 4.6k & 68.9 & 29.5 & 58.3 & 45.3 & 81.1 & 49.3 & 76.3 & 52.4 & 82.9 & 55.6 & 81.1 & 54.9 & 77.6 & 44.8 & 71.9 & 50.9 & 61.3 \\ \hline
        \rowcolor{pink!7} TinySAM \cite{shu2023tinysam} + SA \cite{chen2022adaptformer} + TA \cite{sam2} & 18.7 & 4.6k & 57.0 & 37.0 & 54.9 & 44.8 & 80.9 & 56.6 & 76.6 & 52.5 & 83.4 & 62.1 & 81.2 & 53.6 & 73.8 & 51.9 & 70.9 & 50.3 & 61.7 \\ \hline
        \rowcolor{green!7} TinySAM \cite{shu2023tinysam} + \textbf{SAM-I2V} (Ours) & 18.9 & 4.6k & 73.4 & 37.8 & 55.7 & 48.9 & 84.6 & 61.2 & 77.8 & 58.5 & 86.4 & 65.1 & 83.0 & 59.8 & 81.5 & 54.7 & 72.2 & 55.7 & \textbf{66.0} \\ 
        \hline
        \hline
    \end{tabular}
    \caption{Comparisons of the offline promptable video segmentation with 1-click, 3-click, and 5-click settings on four benchmark datasets (\ie, ES \cite{huang2023neuromorphic}, PU \cite{bekuzarov2023xmem++}, LV \cite{wang2023towards}, and SV \cite{sam2}). ``-'' indicates directly combining existing pre-trained models for inference. ``\textit{Cost}'' is calculated by \textit{GPU number} $\times$ \textit{GPU memory} $\times$ \textit{training hours}. ``\textit{OA}'' denotes the overall segmentation performance. Cutie \cite{cheng2024putting} is excluded from the comparison as it only supports single-frame mask inputs, rendering it unsuitable for offline PVS settings.}
    \label{tab:offline}
\end{table*}

\noindent\textbf{Evaluation Settings.}
To comprehensively assess the capabilities of our SAM-I2V, we perform comparisons on four benchmark datasets: ESD \cite{huang2023neuromorphic} is a high-quality 3D spatial and temporal dataset for object segmentation in an indoor cluttered environment; PUMaVOS \cite{bekuzarov2023xmem++} contains videos with segments around object parts such as a person’s cheek; LV-VIS \cite{wang2023towards} contains videos from a diverse set of open-vocabulary object categories; and SAV-Test \cite{sam2} is the test set of SA-V dataset, consist of 150 videos and 278 masklets.
For these evaluations, we report the standard $J\&F$ metric \cite{pont20172017} which combines region similarity (Jaccard index) and boundary accuracy (F-measure).
We conducted evaluation on the promptable video segmentation. In this task, the model works in an interactive manner, where the point prompts are used to identify the target of interest or to refine the current segmentation result. Following \cite{sam2}, two settings are included in the comparison, namely offline evaluation (PVS-offline) and online evaluation (PVS-online). In both settings, the target is firstly initialized via $N_{\text {click}}$ point prompts. As for PVS-offline, the model segments the target throughout the entire video in $N_{\text {pass}}$ passes, where the interactive frame is selected based on the largest model error (\ie, with lowest IoU between the prediction and ground truth masks) and $N_{\text {click}}$ points are added on such frame to refine the mask in each pass. All the prompts are used to guide the segmentation process in the next pass. PVS-online follows one pass evaluation protocol, where the segmentation process is paused and refined when the result is not reliable (IoU $<$ 0.75 with ground-truth). Also, $N_{\text {click}}$ points are added on the refined frame to generate corrected mask, which is used for subsequent frames. The refine process is repeated $N_{\text {frame}}$ times.

\subsection{Comparison to Existing Methods}
To evaluate the effectiveness of our proposed SAM-I2V, we compared it against seven state-of-the-art methods on both online and offline PVS tasks under different click settings, as reported in Table \ref{tab:online} and \ref{tab:offline}. Besides SAM 2.1 \cite{sam2}, we also include upgradation methods that combine SAM (both base \cite{sam1} and tiny \cite{shu2023tinysam} variants) with existing state-of-the-art video object segmentation models, specifically Xmem++ \cite{bekuzarov2023xmem++} and Cutie \cite{cheng2024putting}, as well as approaches that upgrade SAM by incorporating temporal adaptation (TA) \cite{sam2} and spatial adaptation (SA) \cite{chen2022adaptformer} modules.

\noindent\textbf{Performance Comparison with SAM 2.1.} While SAM 2.1~\cite{sam2} serves as a strong baseline with an overall accuracy of 71.9 (online) and 73.3 (offline), it requires substantial computational resources (2.2 million $G\cdot Hour$). In contrast, our SAM-I2V achieves its 91\% (\ie, 65.3, online) and 90\% (\ie, 66.0, offline) overall scores, with its 0.2\% (\ie, 4.6k $G\cdot Hour$) training cost and its 49\% (\ie, 18.9m vs. 38.9m) model parameters.

\noindent\textbf{Effectiveness over Existing SAM Upgradation Methods.} Compared to SAM~\cite{sam1} + XMem++~\cite{bekuzarov2023xmem++} (157.0M parameters) and TinySAM~\cite{shu2023tinysam} + XMem++ (72.3M parameters), our SAM-I2V offers a substantial reduction in model size (18.9M) while delivering comparable or superior performance. For example, under the online setting, SAM-I2V helps achieve an overall score of 65.3, outperforming SAM + XMem++ (60.4) and SAM + Cutie~\cite{cheng2024putting} (61.7).

\noindent\textbf{Robust Performance Across Interaction Levels.} The online and offline results presented in Table \ref{tab:online} and \ref{tab:offline} reflect that the performance of all methods generally improves as the number of interactions increases. Our SAM-I2V consistently delivers superiority across 1-click, 3-click, and 5-click settings, underscoring its robustness to varying levels of user interaction.

\noindent\textbf{Robust Online and Offline Performance.} SAM-I2V's overall online performance in Table \ref{tab:online} is slightly lower than the offline performance in Table \ref{tab:offline}. This discrepancy arises because, unlike in the offline evaluation, in the online setting the new prompts affect only the frames following the current paused frame and not the preceding frames. Compared to other methods, our approach maintains an advantage in both online and offline settings.

The above results confirm that our SAM-I2V not only matches but often surpasses existing upgradation methods in performance with reduced computational costs, making SAM-I2V a practical choice for PVS tasks.

\subsection{Ablation Study}
We conduct an extensive ablation study to validate the extensiblity of SAM-I2V and the effectiveness of its each key component. \autoref{tab:ie}, \ref{tab:tfi}, \ref{tab:msa}, \ref{tab:mpg} summarize our findings.

\noindent\textbf{Applicability of SAM-I2V to Upgrade Different SAM Models.} In the first ablation study, we investigate the performance of our SAM-I2V to upgrade different SAM models. From the results presented in \autoref{tab:ie}, it is evident that our SAM-I2V effectively upgrades different SAM variants to a PVS model, demonstrate that SAM-I2V is a robust and versatile image-to-video promptable segmentation upgradation approach.

\begin{table}[tbp]
    \centering
    \scriptsize
    \renewcommand{\arraystretch}{1.3}
    \setlength{\tabcolsep}{5pt} 
    \begin{tabular}{l|c|c|c|c|c}
        \hline
        \hline
        Methods & ES\cite{huang2023neuromorphic} & PU\cite{bekuzarov2023xmem++} & LV\cite{wang2023towards} & SV\cite{sam2} & \textit{OA} \\
        \hline
        TinySAM \cite{shu2023tinysam} + TA \cite{sam2} & 80.2 & 45.1 & 72.3 & 49.1 & 61.7 \\
        TinySAM \cite{shu2023tinysam} + SAM-I2V & 84.7 & 60.8 & 73.4 & 58.2 & 69.3 \\
        \hline
        EdgeSAM \cite{zhou2023edgesam} + TA \cite{sam2} & 78.3 & 41.4 & 69.6 & 44.6 & 58.5 \\
        EdgeSAM \cite{zhou2023edgesam} + SAM-I2V & 83.2 & 51.9 & 71.6 & 54.4 & 65.1 \\
        \hline
        \hline
    \end{tabular}
    \caption{Ablation on the applicability of SAM-I2V to upgrade different SAM models for PVS task under the online, 3-click setting.}
    \label{tab:ie}
\end{table}

\noindent\textbf{Effectiveness of the Temporal Feature Integrator.}
As shown in Table~\ref{tab:tfi}, our proposed TFI module outperforms the baseline (TinySAM \cite{shu2023tinysam} + Temporal Associator \cite{sam2}) and other adaptation methods. While previous approaches like ST-Adapter~\cite{pan2022st} and Parallel Adapter~\cite{chen2022adaptformer} perform adaptation within each attention block, our TFI is specifically designed with temporal branch and integration branch to iteratively extract and integrate spatial-temporal features. This allows our model to capture richer temporal dependencies, leading to superior performance.

\begin{table}[tbp]
    \centering
    \scriptsize
    \renewcommand{\arraystretch}{1.3}
    \setlength{\tabcolsep}{5pt} 
    \begin{tabular}{l|c|c|c|c|c}
        \hline
        \hline
        Methods & ES\cite{huang2023neuromorphic} & PU\cite{bekuzarov2023xmem++} & LV\cite{wang2023towards} & SV\cite{sam2} & \textit{OA} \\
        \hline
        Base & 80.2 & 45.1 & 72.3 & 49.1 & 61.7 \\
        \hline
        Base + ST-Adapter \cite{pan2022st} & 79.3 & 54.1 & 73.3 & 50.1 & 64.2 \\
        Base + Parallel Adapter \cite{chen2022adaptformer} & 79.3 & 52.7 & 72.9 & 50.6 & 63.9 \\
        \hline
        Base + TFI & 81.3 & 58.1 & 72.4 & 52.5 & 66.1 \\
        \hline
        \hline
    \end{tabular}
    \caption{Ablation on the effectiveness of the temporal feature integrator TFI for online PVS task under the 3-click setting.}
    \label{tab:tfi}
    \vspace{-3pt}
\end{table}

\noindent\textbf{Effectiveness of the Memory Selective Associator.}
Relying on the previous $z=6$ consecutive frames for segmentation is equivalent to setting $x$ and $y$ in MSA to 0 and 6, respectively. By varying the number of selected global ($x$) and local ($y$) frames in our MSA, we observe that the segmentation performance improves when the MSA selectively retrieves the most relevant frames based on their similarity to the current frame. Specifically, the configuration with $x=3$ global frames and $y=3$ local frames performs better, highlighting the importance of incorporating both short-term and long-term memory. In addition, the ability of MSA to select frames based on relevance rather than uniform sampling allows the model to focus on the most relevant past frames, leading to improved performance.

\noindent\textbf{Effectiveness of the Memory Prompt Generator.}
From Table~\ref{tab:mpg}, we can observe that, (i) introducing MPG (\textit{w/} MPG-1/3/5) consistently improves performance over the one without MPG, demonstrating the effectiveness of the memory-as-prompt strategy in enhancing segmentation accuracy; (ii) varying the number of memory prompt tokens among 1, 3, and 5 yields comparable results, indicating that the MPG is robust to the choice of token quantity within this range; and (iii) removing the Masked Cross-Attention (MCA) mechanism (\ie, \textit{w/} MPG-3 (\textit{w/o} MCA)) leads to a performance drop, highlighting the importance of MCA in focusing on relevant foreground information, eliminating background interference, and thus enhancing the quality of memory prompt generation.

\begin{table}[tbp]
    \centering
    \scriptsize
    \renewcommand{\arraystretch}{1.3}
    \setlength{\tabcolsep}{5.0pt} 
    \begin{tabular}{l|c|c|c|c|c}
        \hline
        \hline
        TinySAM \cite{shu2023tinysam} + SAM-I2V & ES\cite{huang2023neuromorphic} & PU\cite{bekuzarov2023xmem++} & LV\cite{wang2023towards} & SV\cite{sam2} & \textit{OA} \\
        \hline
        \textit{w/o} MSA (\ie, x=0, y=6) & 84.0 & 54.4 & 73.2 & 56.9 & 67.1 \\
        \hline
        \textit{w/} MSA (x=1, y=5) & 84.7 & 54.8 & 73.4 & 57.9 & 67.7 \\
        \textit{w/} MSA (x=2, y=4) & 84.5 & 56.3 & 73.4 & 56.9 & 67.8 \\
        \textit{w/} MSA (x=3, y=3) & 84.7 & 60.8 & 73.4 & 58.2 & 69.3 \\
        \textit{w/} MSA (x=3, y=3) \textit{w/} UFS & 84.1 & 56.0 & 73.0 & 56.5 & 67.4 \\
        \hline
        \hline
    \end{tabular}
    \caption{Ablation on the effectiveness of the memory selective associator MSA for online PVS task under the 3-click setting. ``x'' and ``y'' denote the number of selected global and local frames, respectively. ``UFS'' indicates the uniform frame sampling strategy.}
    \label{tab:msa}
\end{table}

\begin{table}[tbp]
    \centering
    \scriptsize
    \renewcommand{\arraystretch}{1.3}
    \setlength{\tabcolsep}{5pt} 
    \begin{tabular}{l|c|c|c|c|c}
        \hline
        \hline
        TinySAM \cite{shu2023tinysam} + SAM-I2V & ES\cite{huang2023neuromorphic} & PU\cite{bekuzarov2023xmem++} & LV\cite{wang2023towards} & SV\cite{sam2} & \textit{OA} \\
        \hline
        \textit{w/o} MPG & 82.1 & 57.7 & 72.1 & 53.2 & 66.3 \\
        \hline
        \textit{w/} MPG-1 & 84.0 & 58.4 & 73.0 & 55.6 & 67.8 \\
        \textit{w/} MPG-3 & 84.7 & 60.8 & 73.4 & 58.2 & 69.3 \\
        \textit{w/} MPG-5 & 84.5 & 58.6 & 73.6 & 57.1 & 68.5 \\
        \hline
        \textit{w/} MPG-3 (\textit{w/o} MCA) & 84.9 & 58.1 & 72.8 & 55.1 & 67.7 \\
        \hline
        \hline
    \end{tabular}
    \caption{Ablation on the effectiveness of the memory prompt generator MPG for online PVS task under the 3-click setting. ``MPG-\textit{g}'' denotes the MPG with \textit{g} learnable memory prompt tokens. ``MCA'' indicates the masked cross attention.}
    \label{tab:mpg}
\end{table}

\subsection{Limitation and Future Work}
One limitation of our approach is that its accuracy does not well match that of SAM 2.1 \cite{sam2}. Note that the released SAM 2.1 model was trained not only on the SAM-1B and SAV datasets but also on 62.9k proprietary, non-public videos, which our method was unable to utilize. Despite this, our method demonstrates superior performance compared to baseline approaches and creates targeted model design opportunities for downstream applications in resource-constrained academic scenarios. Future work will focus on enhancing our model by incorporating additional publicly available datasets and exploring advanced training techniques to bridge the accuracy gap with SAM 2.1 \cite{sam2}.

\section{Conclusion}
\label{conclusion}
In this work, we introduced SAM-I2V, a training-efficient framework that upgrades the pre-trained Segment Anything Model (SAM) to support promptable video segmentation (PVS). SAM-I2V consists of three key innovations: an image-to-video feature extraction upgrader that extends SAM's static image encoder for spatiotemporal video perception; a memory filtering strategy that selects the most relevant past frames to enhance the utilization of historical information; and a memory-as-prompt mechanism that leverages object memory for temporally consistent mask propagation in dynamic scenes. Comprehensive experiments demonstrate that SAM-I2V achieved over 90\% of SAM 2's performance at only 0.2\% of its training cost. This remarkable training efficiency significantly lowers the barriers for developing diverse PVS models, facilitating rapid innovation and broader practical applications.

\noindent\textbf{Acknowledgments.} This research is supported by the Ministry of Education, Singapore, under its Academic Research Fund Tier 2 (Award No: MOE-T2EP20124-0012).

\clearpage
{
    \small
    \bibliographystyle{ieeenat_fullname}
    \bibliography{main}
}

\clearpage

\section{Appendix}
In the Appendix, we provide additional experimental results and technical details to complement the main paper. Specifically, we demonstrate the applicability of our proposed SAM-I2V across different SAM variants in \autoref{sec:sam_variants}. We also present comprehensive evaluations on the semi-supervised video object segmentation (Semi-VOS) task in \autoref{sec:semi_vos}. Then, we include visual comparisons with state-of-the-art methods in \autoref{sec:visual_comparison}. We further compare and analyze the computational efficiency in \autoref{sec:flops} and explore SAM-I2V's scalability in \autoref{sec:scale}. Finally, detailed descriptions of the modules in our pipeline, including memory encoder, memory attention and mask decoder, are provided in \autoref{sec:technical_details}.

\subsection{Applicability of SAM-I2V Across Different SAM Variants}
\label{sec:sam_variants}
\begin{figure*}
    \centering
    \includegraphics[width=1\linewidth]{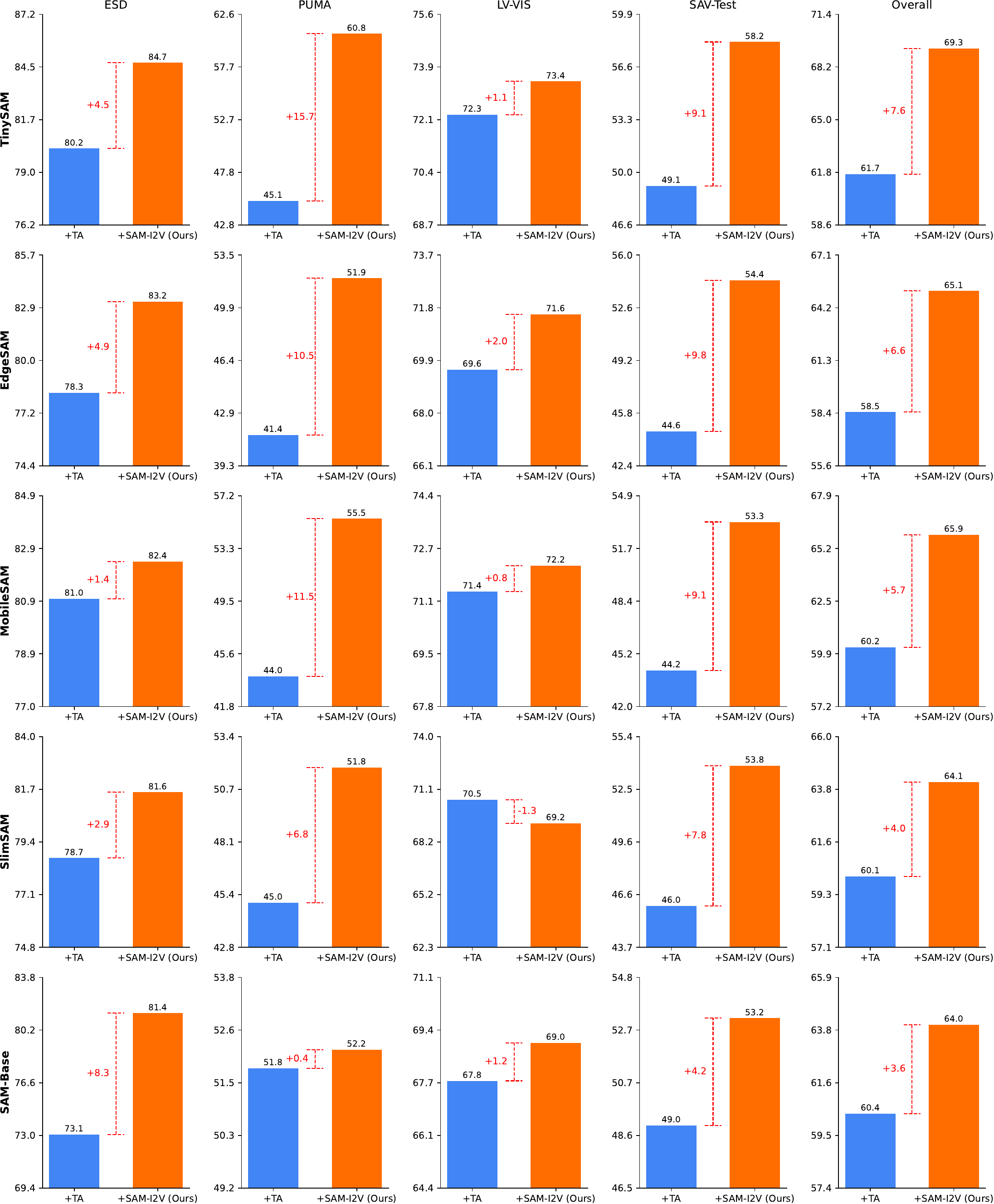}
    \caption{The applicability of our proposed SAM-I2V to upgrade different SAM models for promptable video segmentation (PVS) task under the online, 3-click setting. The comparison includes five SAM variants: TinySAM \cite{shu2023tinysam}, EdgeSAM \cite{zhou2023edgesam}, MobileSAM \cite{zhang2023faster}, SlimSAM \cite{chen20230slimsam}, and SAM-Base \cite{sam1}.}
    \label{fig:boost}
\end{figure*}

To further validate the versatility and robustness of our proposed SAM-I2V approach, we conducted additional experiments on various SAM variants, upgrading them to promptable video segmentation (PVS) models. \autoref{fig:boost} demonstrates the performance improvements (under the online, 3-click PVS setting) when upgraded with SAM-I2V across five SAM variants, including TinySAM \cite{shu2023tinysam}, EdgeSAM \cite{zhou2023edgesam}, MobileSAM \cite{zhang2023faster}, SlimSAM \cite{chen20230slimsam}, and SAM-Base \cite{sam1}. The results highlight the applicability of SAM-I2V in upgrading these promptable image segmentation models to promptable video segmentation models.

For each SAM variant, SAM-I2V consistently brings performance ($J\&F$ \cite{pont20172017}) gains across datasets, including ESD \cite{huang2023neuromorphic}, PUMA \cite{bekuzarov2023xmem++}, LV-VIS \cite{wang2023towards}, and SAV-Test \cite{sam2}, as well as the overall accuracy (\textit{OA}) . For example, in the case of TinySAM \cite{shu2023tinysam}, our approach improves the \textit{OA} from 61.7 to 69.3, representing an improvement of 7.6 points. Similarly, for EdgeSAM \cite{zhou2023edgesam}, SAM-I2V achieves an improvement of 6.6 points in \textit{OA}, demonstrating the robustness of our approach across different SAM models.

These results reaffirm that our SAM-I2V can serve as an efficient and adaptable image-to-video upgradation framework, allowing various SAM models to transition into PVS models without the need for costly training.

\begin{table*}[htbp]
    \centering
    \scriptsize
    \renewcommand{\arraystretch}{1.5}
    \setlength{\tabcolsep}{3.8pt} 
    \begin{tabular}{l|c|c|c|c|c|c|c|c|c|c|c|c|c|c|c|c|c|c|c}
        \hline
        \hline
        \multirow{2}{*}{Methods} & \multirow{2}{*}{\makecell{Para.\\(M)}} & \multirow{2}{*}{Cost} & \multicolumn{4}{c|}{3-click} & \multicolumn{4}{c|}{bounding box} & \multicolumn{4}{c|}{ground-truth mask$^\dagger$} & \multicolumn{4}{c|}{Average} & \multirow{2}{*}{OA} \\ \cline{4-19}
        & & & ES & PU & LV & SV & ES & PU & LV & SV & ES & PU & LV & SV & ES & PU & LV & SV & \\ 
        \hline
        \rowcolor{gray!7} SAM 2.1 \cite{sam2} & 38.9 & 2.2m & 86.6 & 66.8 & 78.1 & 74.4 & 87.8 & 75.4 & 77.7 & 75.1 & 90.2 & 80.9 & 82.2 & 76.5 & 88.2 & 74.4 & 79.3 & 75.3 & 79.3 \\ \hline
        \hline
        \rowcolor{orange!7} SAM \cite{sam1} + XMem++ \cite{bekuzarov2023xmem++} & 157.0 & - & 84.5 & 52.7 & 71.6 & 58.4 & 86.3 & 63.0 & 73.0 & 59.7 & 90.0 & 67.9 & 80.7 & 61.5 & 86.9 & 61.2 & 75.1 & 59.9 & 70.8 \\ \hline
        \rowcolor{orange!7} SAM \cite{sam1} + Cutie \cite{cheng2024putting} & 129.8 & - & 84.6 & 51.0 & 71.9 & 61.1 & 86.1 & 60.0 & 73.4 & 62.8 & 89.7 & 62.5 & 81.4 & 64.8 & 86.8 & 57.8 & 75.6 & 62.9 & 70.8 \\ \hline
        \rowcolor{yellow!7} TinySAM \cite{shu2023tinysam} + XMem++ \cite{bekuzarov2023xmem++} & 72.3 & - & 84.0 & 59.7 & 71.3 & 58.8 & 84.5 & 60.2 & 70.1 & 58.4 & 90.0 & 67.9 & 80.7 & 61.5 & 86.2 & 62.6 & 74.0 & 59.6 & 70.6 \\ \hline
        \rowcolor{yellow!7} TinySAM \cite{shu2023tinysam} + Cutie \cite{cheng2024putting} & 45.1 & - & 84.1 & 56.7 & 72.0 & 61.1 & 84.3 & 58.5 & 70.8 & 61.3 & 89.7 & 62.5 & 81.4 & 64.8 & 86.0 & 59.2 & 74.7 & 62.4 & 70.6 \\
        \hline
        \hline
        \rowcolor{pink!7} TinySAM \cite{shu2023tinysam} + TA \cite{sam2} & 17.8 & 4.6k & 80.0 & 45.0 & 71.2 & 48.6 & 80.2 & 47.3 & 69.3 & 48.6 & 83.1 & 52.5 & 77.3 & 49.6 & 81.1 & 48.3 & 72.6 & 48.9 & 62.7 \\ \hline
        \rowcolor{pink!7} TinySAM \cite{shu2023tinysam} + SA \cite{chen2022adaptformer} + TA \cite{sam2} & 18.7 & 4.6k & 79.0 & 54.9 & 71.5 & 49.8 & 79.9 & 57.6 & 70.5 & 49.8 & 83.5 & 61.1 & 77.4 & 50.5 & 80.8 & 57.9 & 73.1 & 50.0 & 65.5 \\ \hline
        \rowcolor{green!7} TinySAM \cite{shu2023tinysam} + \textbf{SAM-I2V} (Ours) & 18.9 & 4.6k & 84.7 & 58.1 & 72.3 & 57.0 & 83.2 & 66.1 & 69.5 & 56.5 & 87.8 & 68.9 & 79.3 & 59.3 & 85.2 & 64.4 & 73.7 & 57.6 & \textbf{70.2} \\
        \hline
        \hline
    \end{tabular}
    \caption{Comparisons of the semi-supervised video object segmentation with three types of prompt (\ie, 3-click, bounding box, and ground-truth mask) in the first video frame on four benchmark datasets (\ie, ES \cite{huang2023neuromorphic}, PU \cite{bekuzarov2023xmem++}, LV \cite{wang2023towards}, and SV \cite{sam2}). ``-'' indicates directly combining existing pre-trained models for inference. ``\textit{Cost}'' is calculated by \textit{GPU number} $\times$ \textit{GPU memory} $\times$ \textit{training hours}. ``\textit{OA}'' denotes the overall performance. ``$\dagger$'' indicates the case where we directly use masks as inputs into VOS model without using SAM.}
    \label{tab:semi}
\end{table*}

\subsection{Comparison on Semi-Supervised Video Object Segmentation}
\label{sec:semi_vos}

In addition to the online and offline promptable video segmentation results presented in the main paper, we further evaluate our method on the semi-supervised video object segmentation (Semi-VOS) task. For this task, prompts were provided only on the first frame, and the model was tasked with tracking the object through the remainder of the video. This task highlighted the model's capacity for autonomous object tracking without continuous user input, showcasing robustness and generalization in scenarios without ongoing guidance. \autoref{tab:semi} summarizes the performance comparisons under three types of first-frame prompts (\ie, 3-clicks, bounding box, and ground-truth mask) across four benchmark datasets (\ie, ES~\cite{huang2023neuromorphic}, PU~\cite{bekuzarov2023xmem++}, LV~\cite{wang2023towards}, and SV~\cite{sam2}).

Despite not including any VOS datasets during training, our method achieves performance comparable to state-of-the-art VOS methods while utilizing fewer model parameters. Specifically, our method, \textit{TinySAM + SAM-I2V}, attains an overall segmentation performance (\textit{OA}) of \textbf{70.2}, which is competitive with larger models like \textit{TinySAM \cite{shu2023tinysam} + Cutie \cite{cheng2024putting}} (\textit{OA} of 70.6) that require more than double the parameters (45.1M vs. \textbf{18.9}M). Moreover, compared to baseline image-to-video upgrade methods such as \textit{TinySAM + TA~\cite{sam2}} and \textit{TinySAM + SA~\cite{chen2022adaptformer} + TA~\cite{sam2}}, our method shows substantial improvements of \textbf{7.5} and \textbf{4.7} in \textit{OA}, respectively, while maintaining similar computational costs.

Under different first-frame prompt settings, our method consistently outperforms the baselines. For instance, with the \textit{3-click} prompt on the ES \cite{huang2023neuromorphic} dataset, our method achieves an accuracy of \textbf{84.7}, surpassing \textit{TinySAM + TA} (80.0) and \textit{TinySAM + SA + TA} (79.0). Similarly, with the \textit{bounding box} prompt on the PU \cite{bekuzarov2023xmem++} dataset, our method attains \textbf{66.1} accuracy, exceeding the baselines by significant margins. When using the ground-truth mask as the prompt, our method still maintains superior performance across datasets. These results clearly validate the effectiveness of our approach in zero-shot semi-supervised VOS performance under various prompt settings.

\begin{figure*}[t]
    \centering
    \includegraphics[width=.9\linewidth]{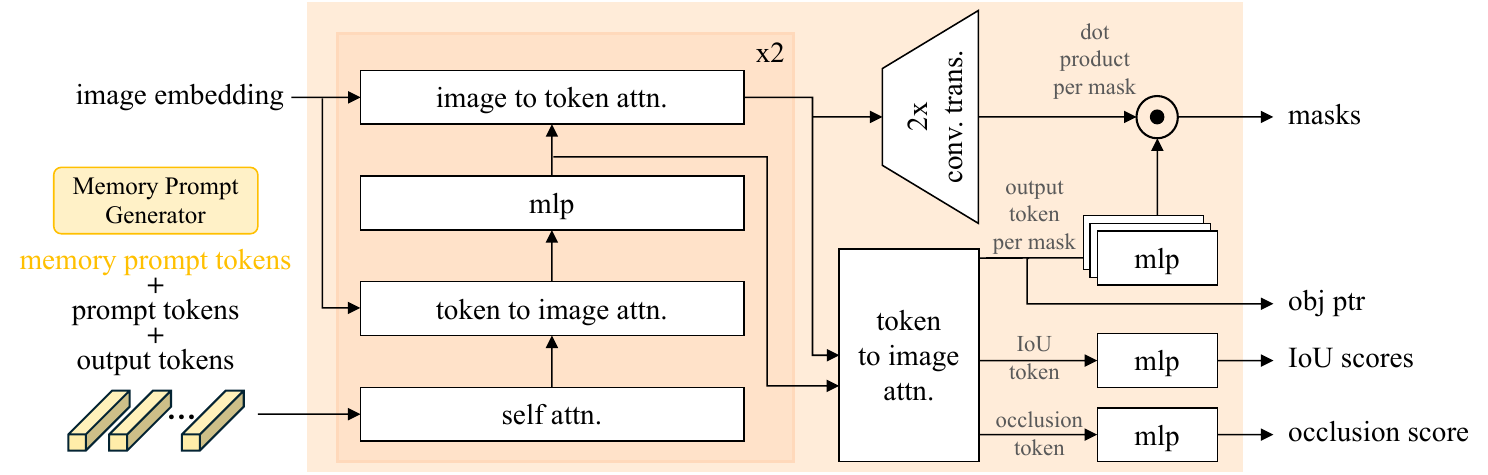}
    \caption{Architecture details of the mask decoder.}
    \label{fig:md}
\end{figure*}

\subsection{Visual Comparison with State-of-the-Art Methods}
\label{sec:visual_comparison}

\begin{figure*}[t]
    \centering
    \includegraphics[width=1\linewidth]{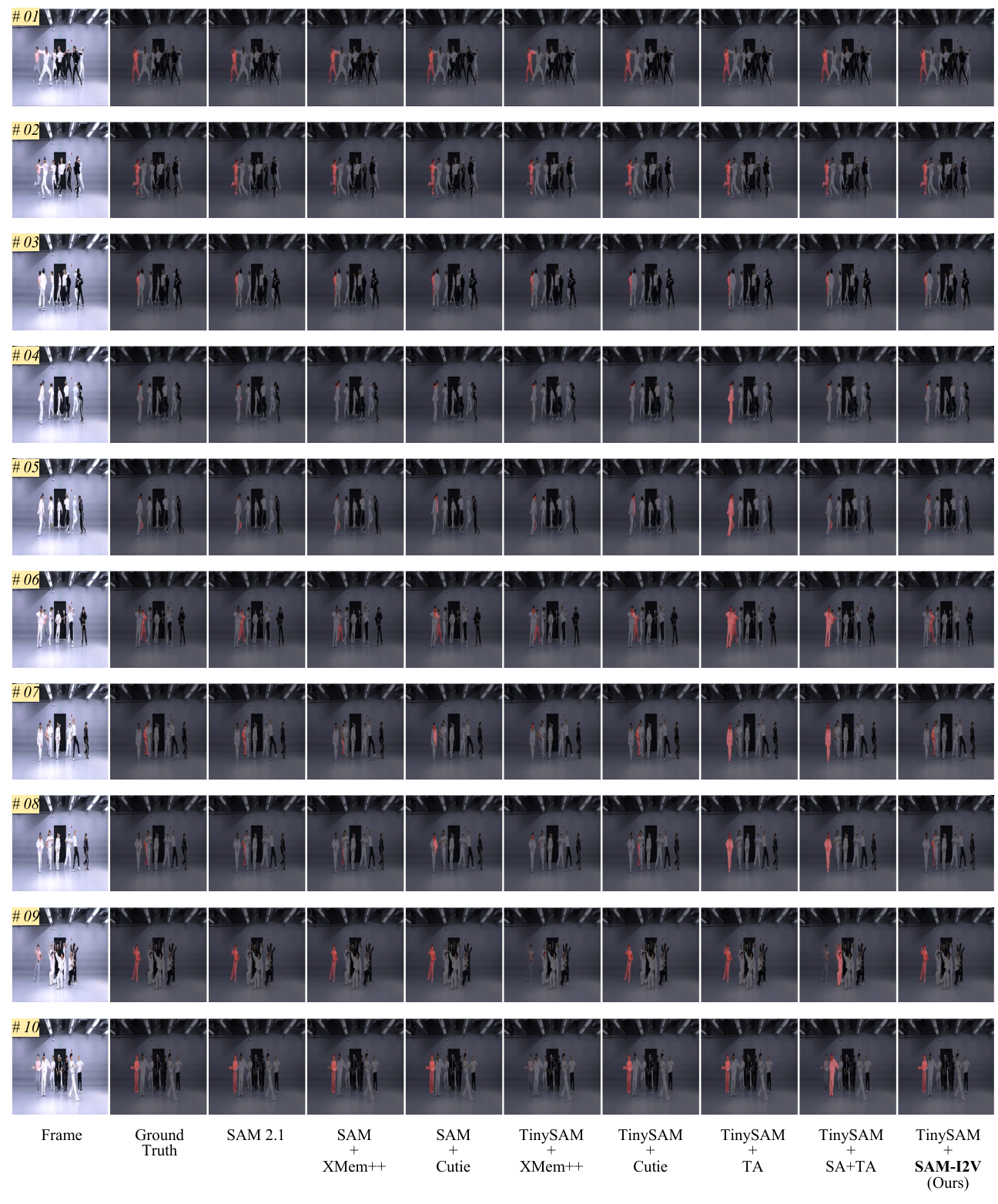}
    \vspace{-20pt}
    \caption{Visual comparison with state-of-the-art methods on a challenging video sequence from PUMaVOS \cite{bekuzarov2023xmem++} dataset. The methods compared include SAM 2.1 \cite{sam2}, SAM \cite{sam1} + XMem++ \cite{bekuzarov2023xmem++}, SAM \cite{sam1} + Cutie \cite{cheng2024putting}, TinySAM \cite{shu2023tinysam} + XMem++ \cite{bekuzarov2023xmem++}, TinySAM \cite{shu2023tinysam} + Cutie \cite{cheng2024putting}, TinySAM \cite{shu2023tinysam} + TA \cite{sam2}, and TinySAM \cite{shu2023tinysam} + SA \cite{chen2022adaptformer} + TA \cite{sam2}. Our method, TinySAM + \textbf{SAM-12V}, demonstrates robust tracking and fine-grained segmentation on video frames, offering competitive performance across challenging scenarios.}
    \label{fig:e1}
\end{figure*}

\begin{figure*}[t]
    \centering
    \includegraphics[width=\textwidth]{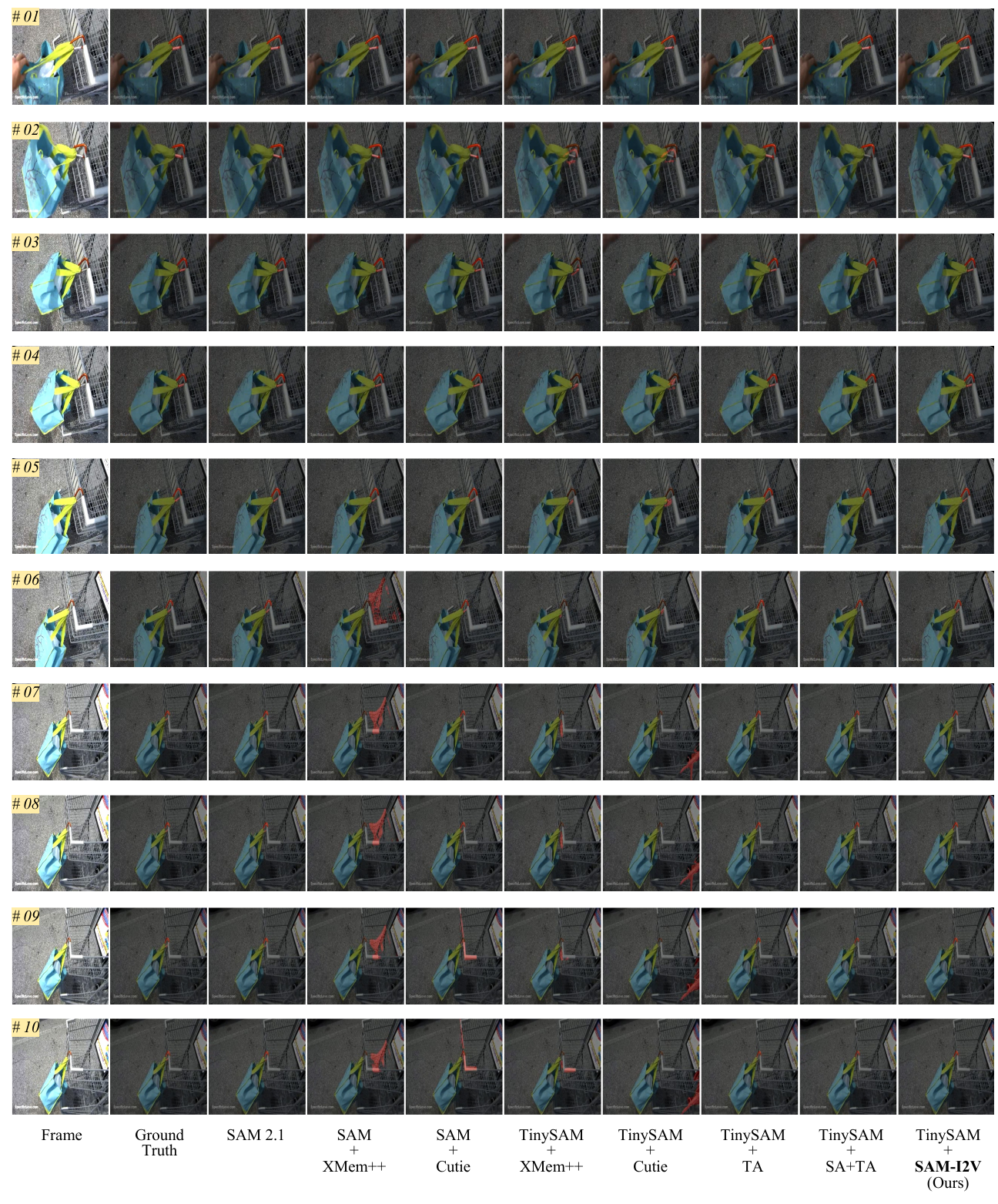}
    \vspace{-20pt}
    \caption{Visual comparison with state-of-the-art methods on a challenging video sequence from LV-VIS \cite{wang2023towards} dataset. The methods compared include SAM 2.1 \cite{sam2}, SAM \cite{sam1} + XMem++ \cite{bekuzarov2023xmem++}, SAM \cite{sam1} + Cutie \cite{cheng2024putting}, TinySAM \cite{shu2023tinysam} + XMem++ \cite{bekuzarov2023xmem++}, TinySAM \cite{shu2023tinysam} + Cutie \cite{cheng2024putting}, TinySAM \cite{shu2023tinysam} + TA \cite{sam2}, and TinySAM \cite{shu2023tinysam} + SA \cite{chen2022adaptformer} + TA \cite{sam2}. Our method, TinySAM + \textbf{SAM-12V}, demonstrates robust tracking and fine-grained segmentation on video frames, offering competitive performance across challenging scenarios.}
    \label{fig:e2}
\end{figure*}

\begin{figure*}[t]
    \centering
    \includegraphics[width=\textwidth]{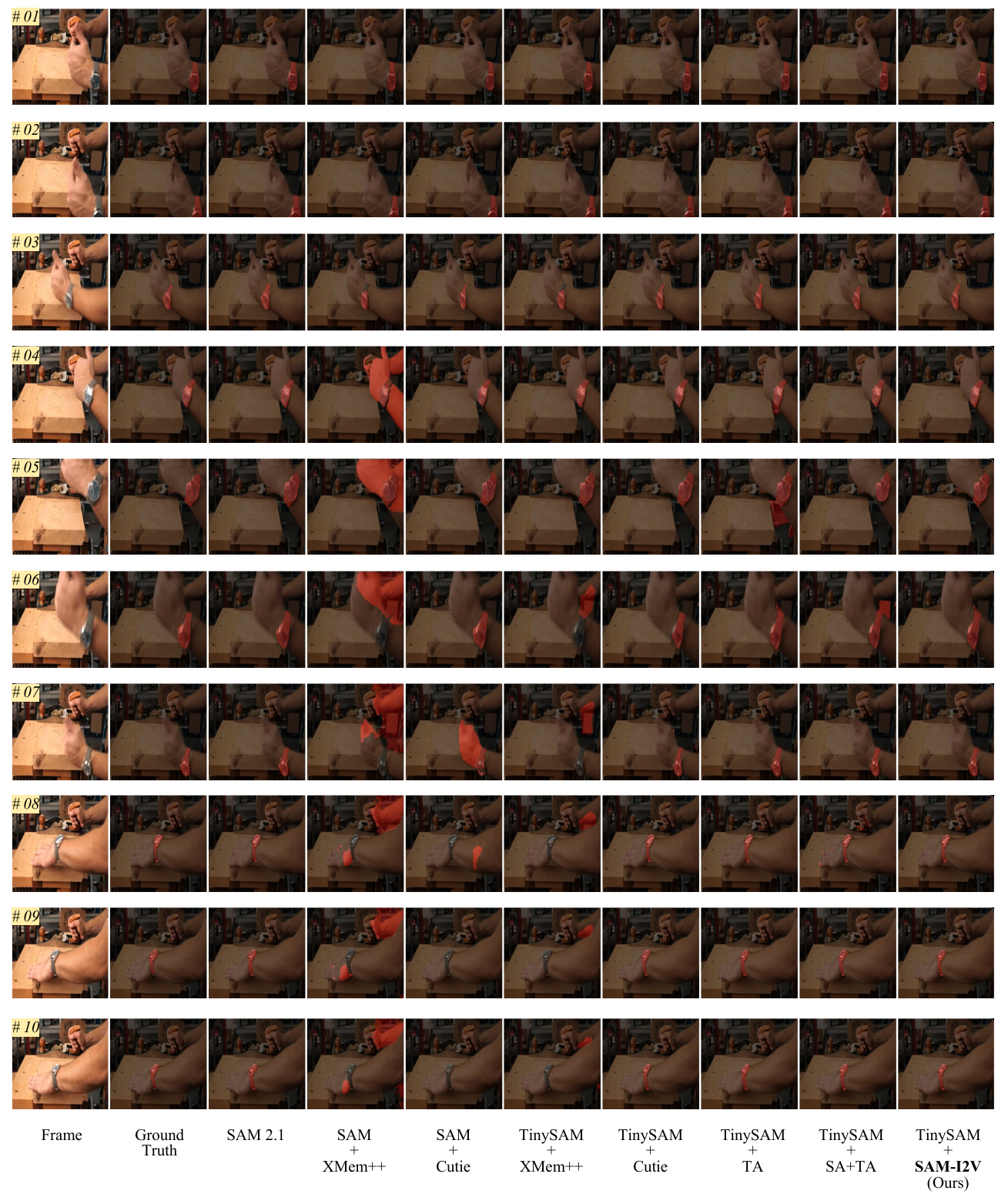}
    \vspace{-20pt}
    \caption{Visual comparison with state-of-the-art methods on a challenging video sequence from LV-VIS \cite{wang2023towards} dataset. The methods compared include SAM 2.1 \cite{sam2}, SAM \cite{sam1} + XMem++ \cite{bekuzarov2023xmem++}, SAM \cite{sam1} + Cutie \cite{cheng2024putting}, TinySAM \cite{shu2023tinysam} + XMem++ \cite{bekuzarov2023xmem++}, TinySAM \cite{shu2023tinysam} + Cutie \cite{cheng2024putting}, TinySAM \cite{shu2023tinysam} + TA \cite{sam2}, and TinySAM \cite{shu2023tinysam} + SA \cite{chen2022adaptformer} + TA \cite{sam2}. Our method, TinySAM + \textbf{SAM-12V}, demonstrates robust tracking and fine-grained segmentation on video frames, offering competitive performance across challenging scenarios.}
    \label{fig:e3}
\end{figure*}

\begin{figure*}[t]
    \centering
    \includegraphics[width=\textwidth]{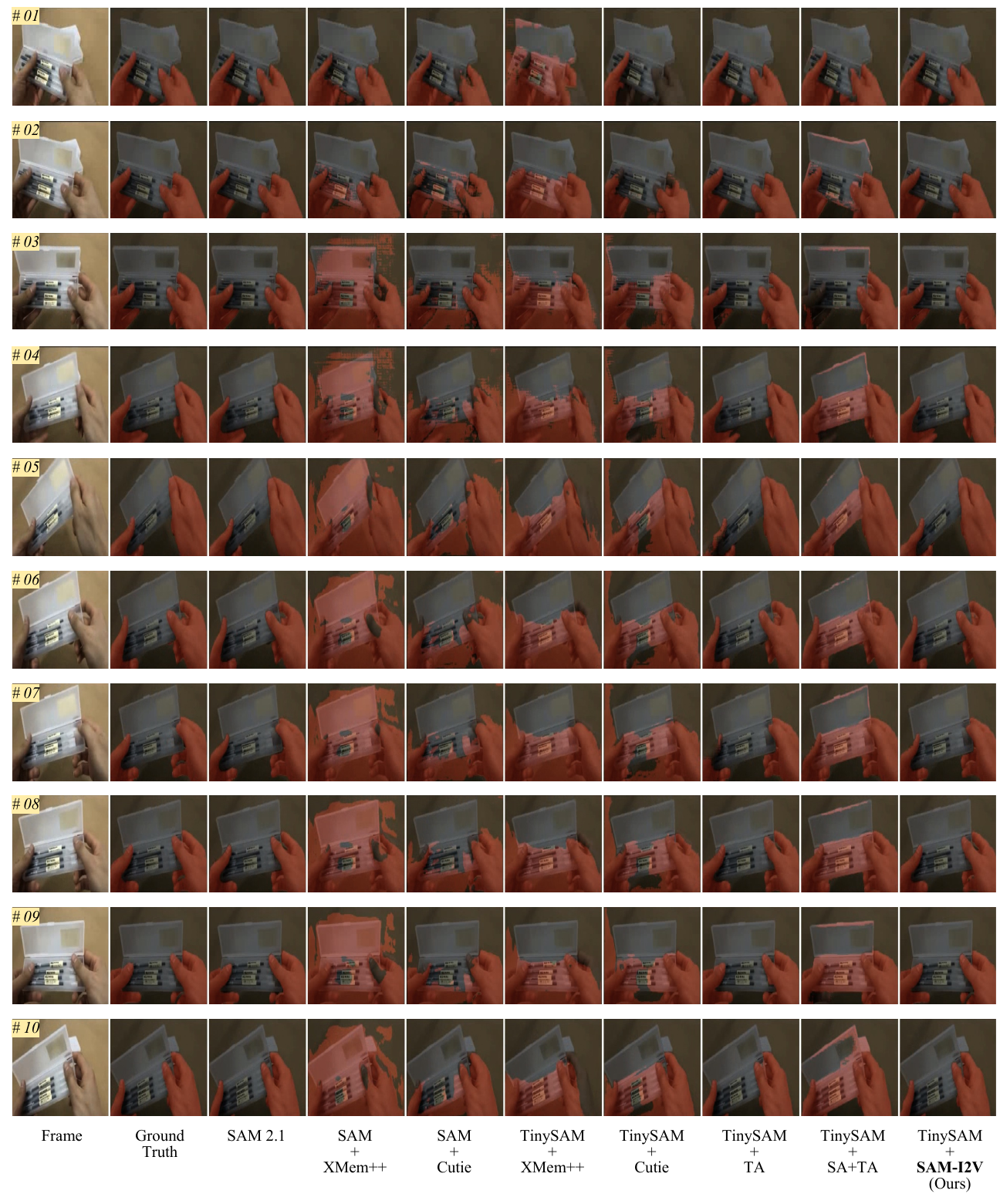}
    \vspace{-20pt}
    \caption{Visual comparison with state-of-the-art methods on a challenging video sequence from LV-VIS \cite{wang2023towards} dataset. The methods compared include SAM 2.1 \cite{sam2}, SAM \cite{sam1} + XMem++ \cite{bekuzarov2023xmem++}, SAM \cite{sam1} + Cutie \cite{cheng2024putting}, TinySAM \cite{shu2023tinysam} + XMem++ \cite{bekuzarov2023xmem++}, TinySAM \cite{shu2023tinysam} + Cutie \cite{cheng2024putting}, TinySAM \cite{shu2023tinysam} + TA \cite{sam2}, and TinySAM \cite{shu2023tinysam} + SA \cite{chen2022adaptformer} + TA \cite{sam2}. Our method, TinySAM + \textbf{SAM-12V}, demonstrates robust tracking and fine-grained segmentation on video frames, offering competitive performance across challenging scenarios.}
    \label{fig:e5}
\end{figure*}

\begin{figure*}[t]
    \centering
    \includegraphics[width=\textwidth]{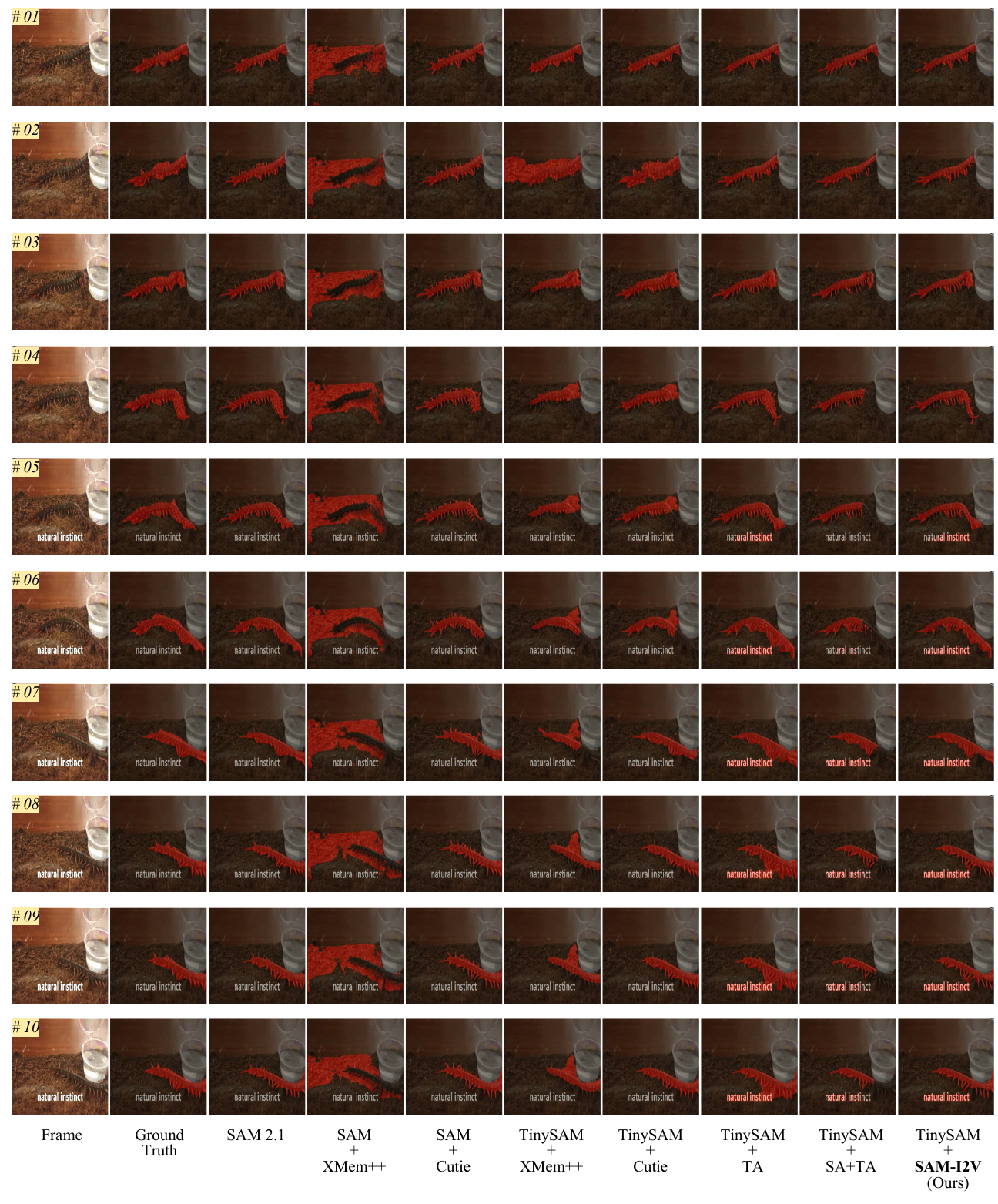}
    \vspace{-20pt}
    \caption{Visual comparison with state-of-the-art methods on a challenging video sequence from LV-VIS \cite{wang2023towards} dataset. The methods compared include SAM 2.1 \cite{sam2}, SAM \cite{sam1} + XMem++ \cite{bekuzarov2023xmem++}, SAM \cite{sam1} + Cutie \cite{cheng2024putting}, TinySAM \cite{shu2023tinysam} + XMem++ \cite{bekuzarov2023xmem++}, TinySAM \cite{shu2023tinysam} + Cutie \cite{cheng2024putting}, TinySAM \cite{shu2023tinysam} + TA \cite{sam2}, and TinySAM \cite{shu2023tinysam} + SA \cite{chen2022adaptformer} + TA \cite{sam2}. Our method, TinySAM + \textbf{SAM-12V}, demonstrates robust tracking and fine-grained segmentation on video frames, offering competitive performance across challenging scenarios.}
    \label{fig:e4}
\end{figure*}

We further present visual comparisons of our method with state-of-the-art approaches under the online, 3-click PVS setting. As shown in Figures~\ref{fig:e1}--\ref{fig:e4}, our method, TinySAM + \textbf{SAM-12V}, robustly tracks and segments objects with fine details across challenging scenarios, maintaining consistency over time. Specifically, \autoref{fig:e1} demonstrates our method's ability to handle \textbf{\textit{occlusion}} effectively; \autoref{fig:e2} showcases accurate segmentation of \textbf{\textit{small objects}}; \autoref{fig:e3} illustrates robust tracking during \textbf{\textit{large spatial movement}s}; \autoref{fig:e5} and \ref{fig:e4} highlight the segmentation of objects with \textbf{\textit{complex shapes}}.
While differences in segmentation results can be observed across methods, our approach consistently delivers competitive performance, particularly in scenarios requiring fine-grained object representation and robust tracking across multiple frames. This clearly demonstrate the effectiveness of our method.

\begin{figure}[t]
    \centering
    \includegraphics[width=1\linewidth]{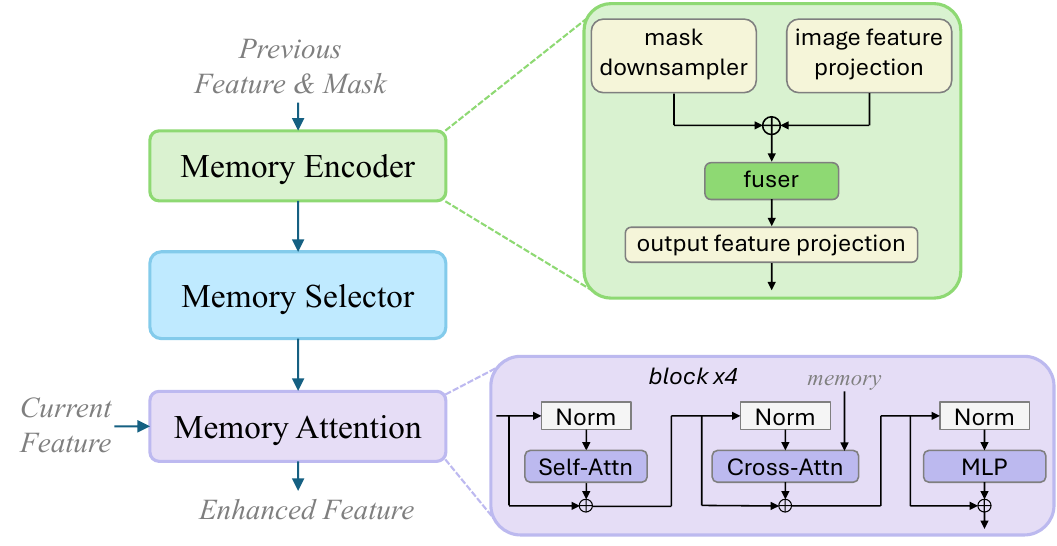}
    \caption{Illustration of the memory selective associator (MSA).}
    \label{fig:msa}
\end{figure}

\subsection{FLOPs Comparison and Analysis}
\label{sec:flops}
We focus in this work on developing an image-to-video SAM-upgrader to support promptable video segmentation with \textbf{\textit{academic-affordable training cost}}. For completeness, we report computational efficiency comparisons in \autoref{tab:flops}.

\begin{table}[tbp]
    \centering
    \scriptsize
    \renewcommand{\arraystretch}{1.3}
    \setlength{\tabcolsep}{13pt}  
    \begin{tabular}{l|c|c}
        \hline
        \hline
        \textbf{Methods} & \textbf{Para. (M)} & \textbf{FLOPs (G)} \\
        \hline
        \rowcolor{gray!7} SAM 2.1 \cite{sam2} & 38.9 & 139.9 \\ \hline
        \rowcolor{orange!7} SAM \cite{sam1} + XMem++ \cite{bekuzarov2023xmem++} & 157.0 & 805.2 \\ \hline
        \rowcolor{orange!7} SAM \cite{sam1} + Cutie \cite{cheng2024putting} & 129.8 & 811.9 \\ \hline
        \rowcolor{yellow!7} TinySAM \cite{shu2023tinysam} + XMem++ \cite{bekuzarov2023xmem++} & 72.3 & 96.3 \\ \hline
        \rowcolor{yellow!7} TinySAM \cite{shu2023tinysam} + Cutie \cite{cheng2024putting} & 45.1 & 103.0 \\ \hline
        \rowcolor{pink!7} TinySAM \cite{shu2023tinysam} + TA \cite{sam2} & 17.8 & 72.9 \\ \hline
        \rowcolor{pink!7} TinySAM \cite{shu2023tinysam} + SA \cite{chen2022adaptformer} + TA \cite{sam2} & 18.7 & 75.0 \\ \hline
        \rowcolor{green!7} TinySAM \cite{shu2023tinysam} + \textbf{SAM-I2V} (Ours) & 18.9 & 87.4 \\
        \hline
        \hline
    \end{tabular}
    \caption{Comparisons in terms of model parameters and FLOPs.}
    \label{tab:flops}
\end{table}

\begin{table}[tbp]
    \centering
    \scriptsize
    \renewcommand{\arraystretch}{1.3}
    \setlength{\tabcolsep}{5.6pt}  
    \begin{tabular}{l|c|c}
        \hline
        \hline
        \textbf{SAM-I2V} & \textbf{Para. (M)} & \textbf{FLOPs (G)} \\
        \hline
        Temporal Feature Integrator & 3.7 & 26.3 \\
        \hline
        Memory Selective Associator: Memory Encoder & 1.4 & 5.8 \\
        Memory Selective Associator: Memory Selector & 0.0 & 0.107  \\
        Memory Selective Associator: Memory Attention & 2.8 & 14.6 \\
        \hline
        Memory Prompt Generator & 0.6 & 0.9 \\
        \hline
        \textbf{Overall} & \textbf{8.5} & \textbf{47.707} \\
        \hline
        \hline
    \end{tabular}
    \caption{Analysis of SAM-I2V's computational efficiency.}
    \label{tab:sam-i2v}
\end{table}

First, we can see from \autoref{tab:flops} that our method exhibits superior computational efficiency (87.4G) compared to SAM 2.1 \cite{sam2} (139.9G) and SAM-based \cite{sam1} variants (805.2G--811.9G), while maintaining parity with TinySAM-driven \cite{shu2023tinysam} approaches (72.9G--103.0G).

Second, as shown in \autoref{tab:sam-i2v}, in our SAM-I2V, the temporal feature integrator (26.3G) and memory selective associator: memory attention (14.6G) collectively account for 85.7\% of the upgrader's computational overhead. This is primarily driven by temporal feature extraction and memory-guided attention operations, respectively, establishing them as critical targets for future FLOPs optimization. Besides, our memory selector (0.107G) achieves extended historical frame association (20 frames versus SAM 2's 6 frames) with extra 158 MB GPU memory footprint. This represents 0.64\% of the total capacity in modern 24GB GPUs, where memory allocation is dominated by similarity score computation between the current frame and the historical feature buffer.

Third, our SAM-I2V achieves image-to-video upgradation with extra 8.5M parameters and 47.707G FLOPs, demonstrating training feasibility under academic GPU constraints. This establishes a practical foundation for training-resource-efficient PVS model development.

\subsection{SAM-I2V's Scalability}
\label{sec:scale}
We further explored the scalability of our SAM-I2V when additional GPU resources are available. As shown in \autoref{tab:scale}, as we increase the training duration (\textit{b}) or the number of GPUs (\textit{c} and \textit{d}), SAM-I2V’s training cost grows proportionally but yields higher SAV-Test performance (59.3 to 65.2), indicating the model’s scalability under greater training investments.

\begin{table}[tbp]
    \centering
    \scriptsize
    \renewcommand{\arraystretch}{1.4}
    \setlength{\tabcolsep}{7pt}  
    \begin{tabular}{c|c|c|c}
        \hline
        \hline
        \textbf{SAM-I2V} & \makecell{\textbf{Training Configuration}\\(\textit{\textbf{\#Num. $\times$ \#Mem. $\times$ \#Dur.}})} & \textbf{Training Cost}  & \makecell{\textbf{SAV-Test}\\(\textit{\textbf{J\&F}})} \\
        \hline
        (\textit{a}) & 8 $\times$ 24 G $\times$ 24 hours & 4.6k & 59.3 \\ \hline
        (\textit{b}) & 8 $\times$ 24 G $\times$ 48 hours & 9.2k & 62.6 \\ \hline
        (\textit{c}) & 16 $\times$ 24 G $\times$ 24 hours & 9.2k & 62.9 \\ \hline
        (\textit{d}) & 32 $\times$ 24 G $\times$ 24 hours & 18.4k & 65.2 \\
        \hline
        \hline
    \end{tabular}
    \caption{Ablation study on SAM-I2V's scalability.}
    \label{tab:scale}
\end{table}

\subsection{Architecture Details}
\label{sec:technical_details}
Here we further present architecture details, expanding on the model description in the main manuscript.

\subsubsection{Memory Selective Associator}
As illustrated in \autoref{fig:msa}, our proposed MSA consists of three sub-networks, \ie, memory encoder, memory selector and memory attention.

The \textbf{memory encoder} is a crucial component designed to transform predictions and image encoder embeddings into representations suitable for future frames in the video segmentation. As shown in the top-right of \autoref{fig:msa}, the memory encoder incorporates a combination of downsampled mask features and projected image features, which are fused using convolutional layers. This fusion process ensures that spatial and contextual information from both input sources is effectively integrated. The resulting fused features are then passed through an output feature projection layer to prepare them for the following memory attention.

The \textbf{memory attention} is designed to condition the current frame features on the past frames' features and predictions. This conditioning is achieved through a stack of $B=4$ transformer blocks, where each block consists of three main components: a self-attention layer, a cross-attention layer, and a feedforward multi-layer perceptron (MLP). The self-attention layer processes the current frame's features to capture intra-frame relationships, ensuring a comprehensive understanding of spatial dependencies within the frame. The cross-attention layer enables interaction between the current frame and the selective memories, which include features from both prompted and unprompted previous frames. The memories are stored in the memory bank and are selectively retrieved based on relevance. Each attention block is normalized before and after the attention operations to maintain stability during training. 2D spatial Rotary Positional Embedding (RoPE) is utilized within self-attention and cross-attention layers to enhance the spatial correspondence of features. Following the attention stages, the MLP further refines the fused features to enhance representational capability. This modular design allows the model to integrate temporal context across video frames, ensuring robust segmentation predictions.

Overall, the memory selective associator plays a pivotal role in enabling the model to maintain and utilize useful temporal information across video frames, facilitating robust mask propogation for accurate video segmentation.

\subsubsection{Mask Decoder}
The mask decoder is designed to segment objects based on image embeddings and prompt tokens. Our architecture extends the design of SAM's mask decoder, incorporating additional memory prompts for enhanced segmentation across video frames.

In \autoref{fig:md}, the prompt tokens represent input guidance such as clicks, bounding boxes, or masks, while memory prompt tokens encode temporal information from previous frames to guide the segmentation in current frame. The inclusion of memory prompt tokens enhances temporal consistency by leveraging the historical context of target objects. The mask decoder employs a sequence of two-way transformer blocks to enable bidirectional attention between image embeddings and tokens. Specifically, the following attention mechanisms are employed:
\begin{compactenum}
    \item \textbf{Self-attention}: Applied to prompt tokens to learn interactions within the token space.
    \item \textbf{Token-to-image attention}: Enables token embeddings to query relevant image features.
    \item \textbf{Image-to-token attention}: Aggregates image feature responses into token representations.
\end{compactenum}

After feature fusion through the transformer blocks, the decoder outputs multiple predictions per frame to handle prompt ambiguities (\eg, a click on the tire of a bicycle could correspond to either the tire or the entire bicycle). To disambiguate these predictions, we propagate only the mask with the highest predicted Intersection over Union (IoU) score. Additionally, following \cite{sam2}, the decoder features an \emph{occlusion prediction head}, implemented as an MLP, which predicts whether the target object is visible in the current frame. This is crucial for handling frames where the object is partially or fully occluded.

\end{document}